\RequirePackage[OT1]{fontenc}
\documentclass[letterpaper, 10 pt, journal, final]{ieeeconf}  
\usepackage[top=54pt, left=54pt, right=54pt, bottom=54pt]{geometry}

\IEEEoverridecommandlockouts                              

\title{\LARGE \bf
Trajectory Optimization and Following for a Three Degrees of Freedom Overactuated Floating Platform
}

\author{A. Bredenbeck$^{1,\,2,\,\dag}$,
S. Vyas$^{3,\,2}$,
M. Zwick$^{2}$,
D. Borrmann$^{1}$,
M.A. Olivares-Mendez$^{4}$,
A. N\"uchter$^{1}$
\thanks{$^{1}$ Informatics VII, University of W\"urzburg, Germany}
\thanks{$^{2}$ Automation and Robotics Group, ESA, Noordwijk, Netherlands}
\thanks{$^{3}$ Robotics Innovation Center (RIC), DFKI Bremen, Germany}
\thanks{$^{4}$ SpaceR-SnT, University of Luxembourg, Luxembourg}
\thanks{$^\dag$ The authors acknowledge the support of Stardust Reloaded project which has received funding from the European Union’s Horizon 2020 research and innovation programme under the Marie Sk\l odowska-Curie grant agreement No 813644
and the Elite Network Bavaria (ENB) for providing funds for the academic degree program ``Satellite Technology''.\vspace{0.5cm}\\
To abide by the \href{https://www.go-fair.org/fair-principles/}{FAIR} principles of science, all software created for this work is available as open source at \href{https://gitlab.com/anton.bredenbeck/ff-trajectories}{\url{gitlab.com/anton.bredenbeck/ff-trajectories}}.}
}

\usepackage[acronyms, nonumberlist]{glossaries}
\usepackage{glossary-longragged}
\makeglossaries 
\newacronym{asdr}{ASDR}{Active Space Debris Removal}
\newacronym{esa}{ESA}{European Space Agency}
\newacronym{leo}{LEO}{Low Earth Orbit}
\newacronym{geo}{GEO}{Geostationary Orbit}
\newacronym{dof}{DoF}{Degrees of Freedom}
\newacronym{rw}{RW}{Reaction-Wheel}
\newacronym{sdf}{SDF}{Simulation Description Format}
\newacronym{acrobat}{ACROBAT}{Air Cushion Robotic Platform}
\newacronym{satsim}{SATSIM}{Satellite Simulator}
\newacronym{recap}{RECAP}{Reaction Control Autonomy Platform}
\newacronym{reacsa}{REACSA}{Reaction Wheel and Satellite Simulator}
\newacronym{lqr}{LQR}{Linear Quadratic Regulator}
\newacronym{mpc}{MPC}{Model Predictive Control}
\newacronym{tvlqr}{TVLQR}{Time-Varying Linear Quadratic Regulator}
\newacronym{awgn}{AWGN}{Additive White Gaussian Noise}
\newacronym{kf}{KF}{Kalman Filter}
\newacronym{ekf}{EKF}{Extended Kalman Filter}
\newacronym{hjb}{HJB}{Hamilton-Jacobi-Bellman}
\newacronym{dre}{DRE}{Differential Riccati Equation}
\newacronym{imu}{IMU}{Inertial Measurement Unit}
\newacronym{minlp}{MINLP}{Mixed-Integer Non-Linear Programming}
\newacronym{nlp}{NLP}{Non-Linear Programming}
\newacronym{pwpf}{PWPF}{Pulse-Width Pulse-Frequency}
\newacronym{pwm}{PWM}{Pulse-Width-Modulation}
\newacronym{moi}{MoI}{Moment of Inertia}
\newacronym{com}{CoM}{Center of Mass}
\newacronym{obc}{OBC}{On-Board Computer}
\newacronym{orl}{ORGL}{Orbital Robotics and GNC Lab}
\newacronym{hil}{HIL}{hardware-in-the-loop}
\newacronym{mocap}{MoCap}{Motion-Capture}
\newacronym{gnc}{GNC}{Guidance, Navigation and Control}
\newacronym{so2}{SO(2)}{Special-Orthogonal Group of Order 2}
\newacronym{bldc}{BLDC}{Brushless Direct-Current}
\newacronym{dem}{DEM}{Digital Elevation Model}
\newacronym{ros2}{ROS2}{Robot-Operating-System 2}

\usepackage{cite}
\usepackage{nicefrac}
\usepackage{amsfonts}
\usepackage[separate-uncertainty = true,multi-part-units=single, tight-spacing=true]{siunitx}
\usepackage{afterpage}
\usepackage[hidelinks]{hyperref}
\usepackage[T1]{fontenc}
\usepackage{mathtools}

\renewcommand{\vec}[1]{\mathbf{#1}}

\begin{document}

\maketitle
\thispagestyle{empty}
\pagestyle{empty}

\begin{abstract}

Space robotics applications, such as \gls{asdr}, require representative testing before launch. 
A commonly used approach to emulate the microgravity environment in space is air-bearing based platforms on flat-floors, such as the European Space Agency's \gls{orl}.
This work proposes a control architecture for a floating platform at the ORGL, equipped with eight solenoid-valve-based thrusters and one reaction wheel. 
The control architecture consists of two main components: a trajectory planner that finds optimal trajectories connecting two states and a trajectory follower that follows any physically feasible trajectory.
The controller is first evaluated within an introduced simulation, achieving a 100\% success rate at finding and following trajectories to the origin within a Monte-Carlo test.
Individual trajectories are also successfully followed by the physical system. 
In this work, we showcase the ability of the controller to reject disturbances and follow a straight-line trajectory within tens of centimeters.
\end{abstract}

\vspace{-0.15cm}
\section{INTRODUCTION}

Space debris is widely recognized as a significant challenge to all future activities in space~\cite{Crowther1241, NEWMAN201830,Kessler1978}.
In the extreme case, it could entirely prohibit the possibility of safe spaceflight.
Becoming known as the ``Kessler-Syndrome'', the described effect gained traction in recent years for multiple reasons. 
Firstly, today a significant portion of debris is caused by human activity. 
Some of the most significant contributions stem from anti-satellite missile tests and satellite collisions~\cite{Neuneck2008, Kelso2009AnalysisOT}.
Despite stringent requirements on de-orbiting satellites, many community members argue that these are insufficient to avoid the Kessler-Syndrome mentioned above and campaign for \gls{asdr}~\cite{Chatterjee2015,MARK2019194}.

\begin{figure}
	\centering
	\includegraphics[width=0.49\textwidth]{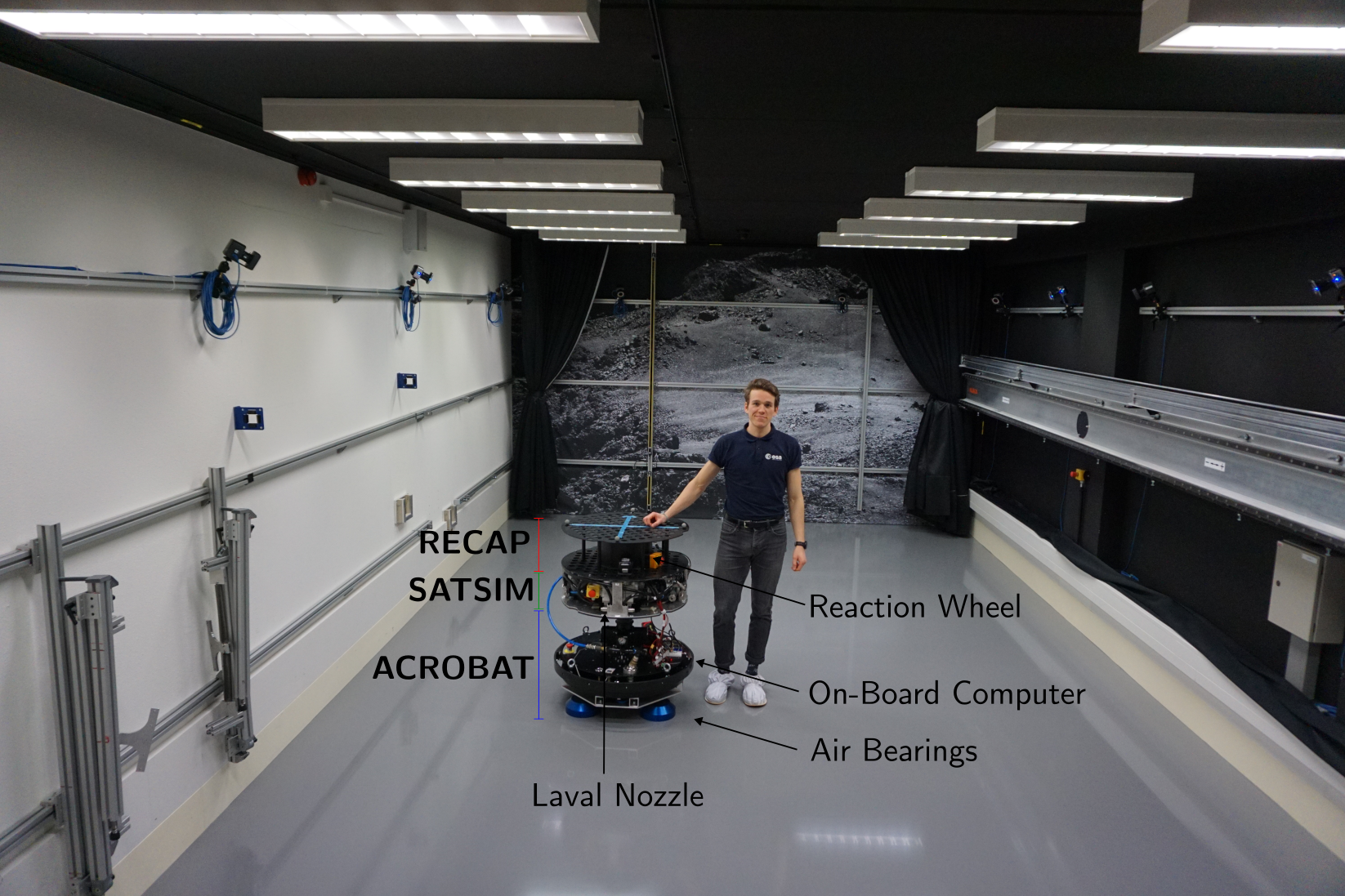}
	\caption{The first author at the \gls{orl} at ESTEC. An epoxy flat-floor for floating air-bearing platforms, where slight unevenness induces disturbances.}
	\label{fig:orl}
\end{figure} 

Given the proximity of the servicing/capturing device and the respective client, the missions carry above-average risk. 
Hence, guaranteeing flawless operation as well as possible prior to launch is highly prioritized. 
For this purpose, ground-test facilities that provide system-level evaluation are necessary. 
The servicing/capturing system will eventually operate in a zero-g environment while all ground facilities are subject to Earth's gravity. 
Hence, precisely replicating the entire operating domain proves to be very difficult.  

Air-bearing-based platforms have become the most popular testing facility that simulates microgravity, with many examples in use~\cite{10.1007/3-540-45118-8_22,ABF,Papadopoulos2008,redah2019,Wehrmann2017,RYBUS2016239}.

Figure~\ref{fig:orl} depicts one of these facilities: the \gls{orl} at ESTEC, a \gls{esa} facility.
One of the testing platforms at the \gls{orl} provides a realistic actuator assembly, with several cold gas thrusters and a reaction wheel. 
This system functions as a base platform for testing new technologies or as a dummy target for capturing tests. 
Thus it is of interest to control the system along desired trajectories as a target or to enable the movement of the tested technology.
For all air-bearing-based floating platforms, the most limiting factor for the test duration is the amount of cold gas stored onboard, which provides the necessary pressure to keep the platform floating. 
The compressed breathing air also functions as the propellant for the thrusters that control the position and orientation (pose) of the system. 
Therefore, a controller aiming to prolong the test duration must use the thrusters in a propellant-optimal manner along any trajectory.
This objective is similarly vital for the control of satellites, as depletion of fuel limits the lifetime of a satellite.

The main contributions of this work are the open-source implementation and validation of a trajectory finding and following controller for free-floating systems, which are subjected to binary actuation constraints.
It proposes to find optimal trajectories between two points in state-space and then use optimal control methods combined with a modulation scheme for the thrusters to follow the desired trajectory. 

The remainder of the document is structured as follows:
After introducing the system model in section~\ref{sec:SystemModel}, section~\ref{sec:controller} gives an overview of the control architecture. 
Section~\ref{sec:results} carries out an evaluation in simulation and on the physical system before finally summarizing and discussing the results in section~\ref{sec:conclusion}.
\vspace{-0.2cm}
\section{SYSTEM MODEL}\label{sec:SystemModel}

The used platform is the combination of three previously existing, modular platforms: \gls{acrobat}, \gls{satsim}, and \gls{recap}~\cite{orgl}.
The platforms are stacked to form the overall system, and each provides a different functionality:
\begin{itemize}
  \item \gls{acrobat} provides the base of the platform. 
  Using three New Way \SI{200}{\milli\meter} air-bearings which are supplied with compressed air at \SI{4.8}{\bar} it enables the micro-gravity behavior on the flat ground. 
  \item \gls{satsim} provides air tanks that supply the \gls{acrobat} air bearing as well as the thrusters that \gls{satsim} uses to provide thrust to the stack.
  \gls{satsim} combines eight thrusters that are arranged pairwise on each side of the platform to induce forces parallel to the coordinate axes of the local robot coordinate system as indicated in Figure~\ref{fig:reacsa}
  \item \gls{recap} provides the \gls{rw} used for yaw control.
\end{itemize}  
Their mass, \gls{moi}, and size properties are shown in Table~\ref{tab:mass_and_size_properties}.
\begin{table}
  \centering
  \vspace{0.15cm}
  \caption{Physical properties of the system.}
  \begin{tabular}{@{}lllll@{}}\hline
  Subsystem & Mass &\gls{moi}& Height & Radius\\\hline\hline
  ACROBAT & \SI{154}{\kilo\gram}& \SI{10.090}{\kilo\gram\meter^2} &\SI{62.5}{\centi\meter}&\SI{35}{\centi\meter}\\\hline
  SATSIM & \SI{50}{\kilo\gram}& \SI{1.416}{\kilo\gram\meter^2} &\SI{20}{\centi\meter}&\SI{35}{\centi\meter}\\\hline
  RECAP & \SI{13.66}{\kilo\gram}& \SI{0.67}{\kilo\gram\meter^2} &\SI{20}{\centi\meter}&\SI{35}{\centi\meter}\\\hline
  RW & \SI{4.01}{\kilo\gram}& \SI{0.047}{\kilo\gram\meter^2}& -- & -- \\\hline\hline
  $\Sigma$ & \SI{221.67}{\kilo\gram}&\SI{12.223}{\kilo\gram\meter^2} & \SI{102.5}{\centi\meter}& -- \\\hline
  \end{tabular}
  \label{tab:mass_and_size_properties}
\end{table}

This section characterizes the overall system. 
In particular, this section identifies the thrust by an individual thruster and the motion model.

\vspace{-0.1cm}
\subsection{Thruster Characterization} 

The thrusters built into this system are a combination of a tank holding pressurized gas, an intermediate pressure regulator to achieve a lower operating pressure, a regulating solenoid valve, and a Laval nozzle through which the gas escapes.
Solenoid valves only allow for the states on and off.
Thus, their nominal force, when opened, is of interest. 
At a regulated operating pressure of $\SI{5.0(2)}{\bar}$ the force is measured to be $\SI{10.0(7)}{\newton}$.
The thrusters can hold this peak force for pulses of \SI{100}{\milli\second} but experience a significant drop for longer pulses in the order of seconds. 
In the following, this work assumes that the force is constant for the opening duration, and no pulses long enough to induce significant deviation from this model are considered.  

\subsection{Dynamic Model}

\begin{figure}
  \centering
  \vspace{0.15cm}
  \includegraphics[width=0.49\textwidth, trim={0cm 10.5cm 1cm 1cm}, clip]{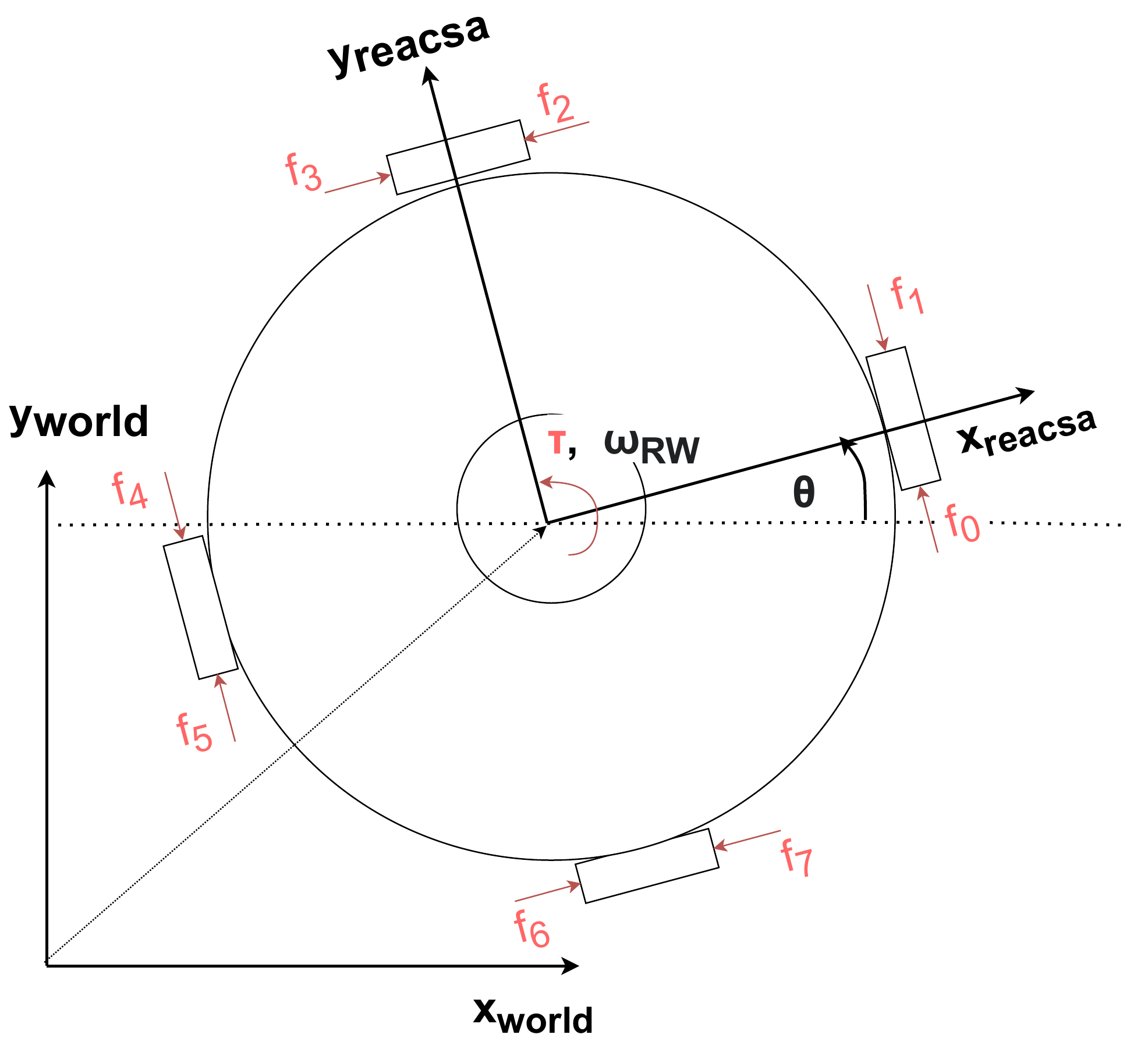}
  \caption{Definitions of coordinate system and wrenches imposed by the actuators.}
  \label{fig:reacsa}  
\end{figure}

The following derives the overall motion model using the thrust identified above and a torque-based model for the \gls{rw}.
First, we define all actuators as in Figure~\ref{fig:reacsa}.
The position $(x,y)$ of the system is defined relative to some world coordinate system, and the orientation $\theta$ is defined relative to that $x$ axis. 

\noindent Using the state and control vectors:
\begin{align}
  \vec{x} &= \begin{bmatrix}x & y & \theta & \dot{x} & \dot{y} & \dot{\theta} & \omega_{RW}\end{bmatrix}^T\\
    \vec{u} &= \begin{bmatrix}\tau & f_0& f_1& f_2& f_3& f_4& f_5& f_6& f_7\end{bmatrix}^T
\end{align}  
with the forces of the thrusters $f_0$ to $f_7$, the velocity of the reaction wheel $\omega_{RW}$ and the torque $\tau$ on the \gls{rw}, the resulting continuous state equation is: 
\begin{align}
  \begin{aligned}        
    &\dot{\vec{x}} = \vec{f}\left(\vec{x},\vec{u}\right) = \begin{bmatrix}
    \vec{0}^{3\times3} & \vec{I}^{3\times3} & 0\\
    & \vec{0}^{4\times7}
    \end{bmatrix} \vec{x} 
    \\&+\resizebox{0.4\textwidth}{!}{
    $\begin{bmatrix}&&&& \vec{0}^{3\times7}&&&\\
    0 & \frac{-s_\theta}{m} & \frac{s_\theta}{m} & \frac{-c_\theta}{m} & \frac{c_\theta}{m} & \frac{s_\theta}{m} & \frac{-s_\theta}{m} & \frac{c_\theta}{m} & \frac{-c_\theta}{m}\\
    0 & \frac{c_\theta}{m} & \frac{-c_\theta}{m} & \frac{-s_\theta}{m} & \frac{s_\theta}{m} & \frac{-c_\theta}{m} & \frac{c_\theta}{m} & \frac{s_\theta}{m} & \frac{-s_\theta}{m}\\
    \frac{-1}{I_b} & \frac{r}{I_b} & \frac{-r}{I_b}& \frac{r}{I_b} & \frac{-r}{I_b} & \frac{r}{I_b} & \frac{-r}{I_b} & \frac{r}{I_b} & \frac{-r}{I_b}\\
    \frac{1}{I_w} &&&& \vec{0}^{1\times6}&&\end{bmatrix}\vec{u}$}\;\;,
  \end{aligned}\label{eq:continuousDynamics}
\end{align} 
where $s_\theta$ and $c_\theta$ denote the sine and cosine of the respective angle, $m$ is the system mass, $r$ the radius of the cylindrical body, $I_w$ and $I_b$ are the \gls{moi} of the \gls{rw} and the overall system respectively. 

\vspace{-0.15cm}
\section{CONTROLLER}\label{sec:controller}

\subsection{Overview}

\begin{figure}
  \centering
	\includegraphics[width=0.49\textwidth, trim={0 11.05cm 0cm 0cm}, clip]{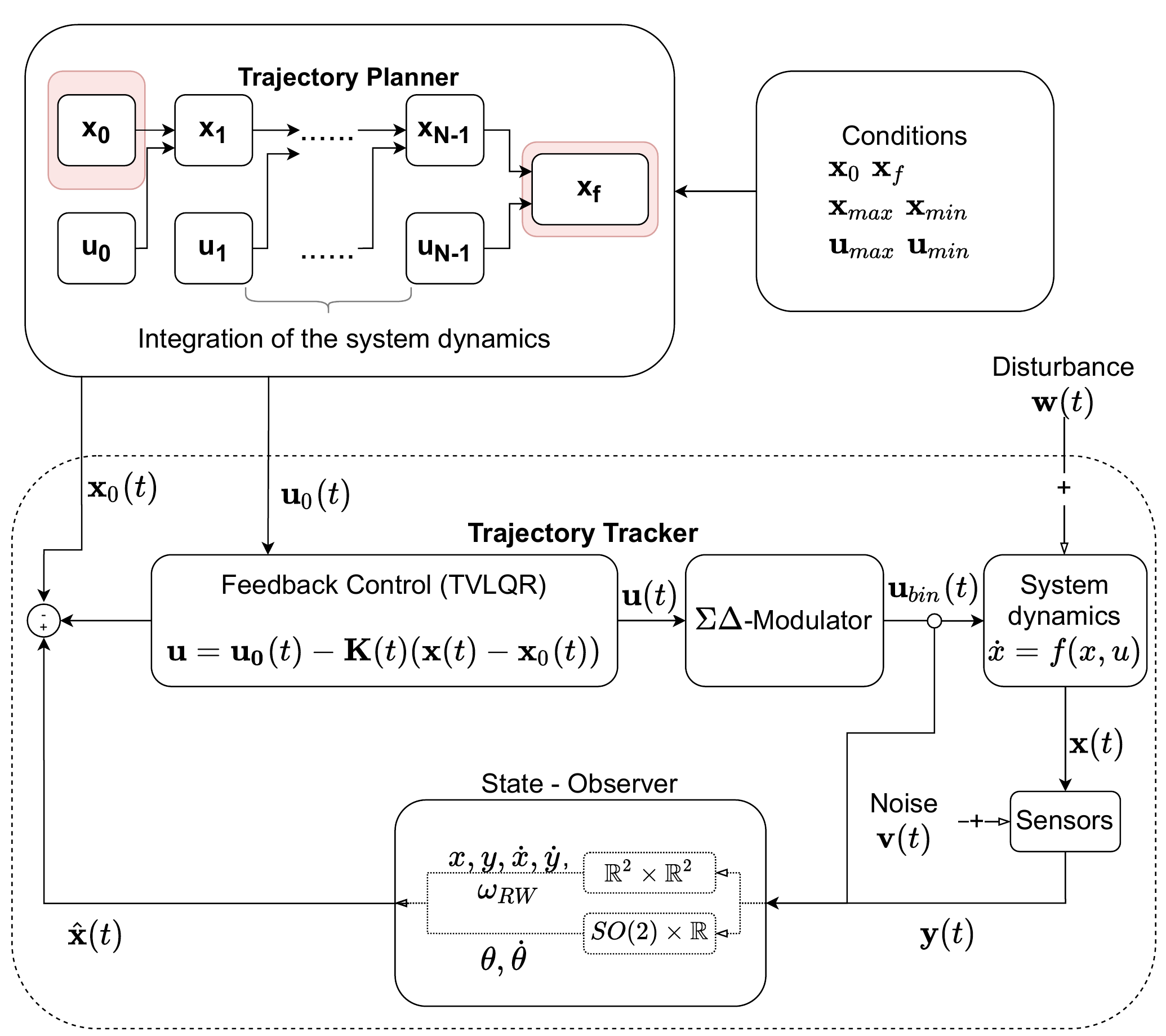}
	\caption{Block diagram of the control architecture, consisting of a trajectory planner and follower (\gls{tvlqr}, a $\Sigma\Delta$-Modulator, and an Observer).}
	\label{fig:controlstructure}
\end{figure}

The block diagram in Figure~\ref{fig:controlstructure} gives an overview of the entire control architecture.
The system consists of two main modules: the trajectory planner and the trajectory tracker. 
The planner computes an optimal trajectory a-priori, given system constraints specified by the user. 
The resulting trajectory has the form of $N$ knot-points, each comprised of a state and a control vector, which in combination satisfy the dynamics of the system and the specified constraints if physically feasible.
The trajectory tracker then tries to follow the trajectory using three sub-modules: the continuous feedback controller, the modulator, and the observer. 
The feedback controller computes the optimal, continuous force required to follow the trajectory, using \gls{tvlqr}~\cite{bertsekas2012dynamic,Underactuated}.
The modulator then chooses opening times for the on/off thrusters to match the continuous force best using a $\Sigma\Delta$-Modulation scheme~\cite{Zappulla2017ExperimentalEM}.
A motor encoder and a \gls{mocap} system provide measurements of the \gls{rw} velocity and the system pose, respectively.
The observer uses the most recent measurements available at the sensors and the commanded control input to optimally estimate the current system state.
The state estimates and the torque commands are computed at \SI{100}{\hertz} while the force command is calculated at a slower \SI{10}{\hertz} to better abide by the physical limitations of the thrusters.
Further, the trajectory follower can follow the previously computed optimal trajectory and any other admissible (physically feasible) trajectory. 

\subsection{Trajectory Planner: Optimizing over Discrete States and Control Inputs}
\subsubsection{Cost Function}
The cost function $J$ is the main factor determining the shape of the resulting trajectory. 
The two qualitative criteria that should be optimized are propellant efficiency and trajectory duration. 
Given the actuator limits, there is a minimal time for the system to reach the final state.
This trajectory is called time-optimal.
On the other hand, if the desired finishing time is infinite, the most propellant-efficient action is to do nothing.
In order to find a trajectory that satisfies both criteria, the approach is as follows:

First, find the time-optimal trajectory using the cost function $J = t_f$. 
In addition, it is enforced that $t_f \geq 0$.
It is well known that the time-optimal solution for such systems is a ``bang-bang'' controller~\cite{betts2020practical} -- a controller jumping between maximal and minimal control values.
However, it also provides a lower bound for the overall time required to follow the trajectory.

Afterward, to define a desired final time, the time-optimal final time is multiplied with a buffer factor $\alpha$: $t_{f,des} = \alpha \cdot t_f^*$, where the buffer factor $\alpha$ is chosen heuristically to provide slow, smooth movement. 
The number of tunable heuristics is the main difference between this approach and simultaneously optimizing the final time and actuation costs. 
Simultaneous optimization requires two separate weights for both cost functions, whereas the separation reduces this to one.

The goal is to find the trajectory that minimizes the propellant usage, finishing at the desired time. 
The force exerted by the thrusters is approximately proportional to the propellant used; therefore, a reasonable proxy for minimal-propellant is minimal force.
The cost function is then
\begin{align}
	J = \sum_{k=1}^N\vec{u}_k\vec{R}\vec{u}_k^T\;\;,
\end{align}
where $\vec{R}$ is a diagonal matrix that contains the respective weights for each control value. 
By choosing a large weight for all thrusters and a small weight for the \gls{rw} the actuation of the thrusters is minimized.

\subsubsection{Direct Collocation}
To find an optimal trajectory that minimizes the above cost functions, we use direct collocation~\cite{hargraves1987,kelly2017}.
Its key aspect is to discretize the trajectory at $N$ instances of time $t_k$, which are denoted as knot-points. 
Each knot-point is subject to constraints that reflect the maximal and minimal state and control values and constraints concerning its neighbors. 
In particular, the dynamic model (as derived in section~\ref{sec:SystemModel}) must be satisfied. 
Since that model is continuous, we use  Hermite-Simpson collocation~\cite{hargraves1987} to integrate its dynamics.
Then it is ensured that the system dynamics are satisfied for any set of subsequent knot-points. 
In other words, the state and the control applied at some knot-point must result in the state at the next knot-point according to the integrated dynamics.
From this it is possible to derive a minimization problem on all $N\times\left(\text{Dim}(\vec{x})+\text{Dim}(\vec{u})\right)$  variables. 
For the time-optimal problem, the final time adds one more decision variable.
However, the optimization problem that incorporates the binary restriction on the actuators is of the class \gls{minlp}. 
These problems are increasingly difficult to solve because they are NP-hard and combine challenges of handling non-linearities with a combinatorial explosion of integer variables~\cite{belotti_2013}.
By relaxing the binary condition on the respective control variables and assuming them to be continuous, the problem reduces to a problem of the class \gls{nlp}, which can be solved significantly faster.
The resulting optimization problem over all states $X$ and control values $U$ at all knot-points is:
\begin{align}
\begin{aligned}
	& \min_{X, U} \left\{J(X,U)\right\}\;\;\;\;\forall k \in [0, N-1]\text{ s.t. } \\
	&\vec{x}(0) = \vec{x}_{init}, \;\;\; \vec{x}(t_f) = \vec{x}_{final}\\ 
	&\vec{x}_{min} \leq \vec{x}_{k} \leq \vec{x}_{max}, \quad \vec{u}_{min} \leq \vec{u}_{k} \leq \vec{u}_{max}\\
	&\vec{x}_{k+1} - \vec{x}_{k} = \frac{\Delta t}{6}(\vec{f}_k + 4 \vec{f}_{k+1/2} + \vec{f}_{k+1})\\
	& \text{where }	\vec{x}_{k+1/2} = \frac{1}{2}(\vec{x}_{k} + \vec{x}_{k+1}) + \frac{\Delta t}{8}(\vec{f}_{k} - \vec{f}_{k+1})\\
	& \text{and } \vec{u}_{k+1/2} = \frac{1}{2}(\vec{u}_{k} + \vec{u}_{k+1})\\
\end{aligned}
\label{eq:cont_opt_prob}
\end{align}
Readily available, open-source solvers, such as IPOPT (Interior Point OPTimiser)~\cite{Wchter2006OnTI}, using the programming interface provided by Drake~\cite{drake}, find a solution for a zero initial-and final velocity trajectory within ten seconds. 
However, this relaxation yields trajectories that demand continuous control input which the discrete or binary actuator cannot provide. 
Therefore, any trajectory tracking controller needs to consider how to translate the desired control input into the discrete space.
This is further discussed in subsection~\ref{subsec:trajfollower}.

\subsection{Trajectory Follower}\label{subsec:trajfollower}

\subsubsection{Time Varying Feedback Controller}\label{subsec:trajtracker}

Once there is an optimal trajectory to be followed, the robot should find the controls that, given the current state, propels the robot along the trajectory.
We propose to use a full state feedback controller based on the \gls{tvlqr} formulation.
Solving the \gls{dre} by initializing the cost-to-go with its value at the final time and integrating it backward in time yields the respective cost-to-go matrices at different times $t$ and from this, one can compute the feedback gain matrices $\vec{K}(t)$~\cite{Underactuated}. 
The resulting control law is:
\begin{align}
	\vec{u}(t) = \vec{u_0}(t) + \vec{K}(t)(\vec{x}(t) - \vec{x}_0(t)),
\end{align}
where $\vec{u}(t)$ is the desired control input, $\vec{u_0}(t)$ is the feed-forward control from the pre-computed trajectory, $\vec{x}(t)$ is the current state, and $\vec{x}_0(t)$ is the desired state according to the trajectory.
Note that solving the \gls{dre} backward in time implies the controller must pre-compute the gain matrices for a trajectory beforehand, resulting in a few additional seconds of computation time during the initialization. 

\subsubsection{Sigma-Delta-Modulator}

Given that there are discrete actuators in the present system, one needs to answer how to modulate the continuous control signal derived in the previous section.
One approach introduced in~\cite{Zappulla2017ExperimentalEM} is using a $\Sigma\Delta$-Modulator, a technique commonly used in analog to digital modulation, to modulate the continuous force onto the binary actuators. 
The basic concept of a $\Sigma\Delta$-Modulator is to trigger a pulse as soon as the integrated error between the desired and current output reaches some specific threshold.
The general concept of the $\Sigma\Delta$-Modulator is the following:
\begin{enumerate}
 	\item Sample the continuous signal $u(t)$ at some frequency $f_{smpl}$ using sample and hold.
 	\item Compute the error $e_{\Sigma\Delta}(t)$ by taking the difference of the current thruster output $y_{Thrust}(t)$ (which is already modulated) to the desired value.
 	\item Integrate the error by numeric integration where $\Delta t$ is the time difference between two samples:\\
 	$w_e(t) = \int_{t_{0}}^t e_{\Sigma\Delta}(\tau) d\tau \approx \Delta t\sum_{i = 0}^T e_{\Sigma\Delta}(i\cdot\Delta t)$ .
 	\item Once the integrator value surpasses some threshold, i.e. $w_e(t) > \epsilon$, trigger a pulse. 
 	We guarantee equivalence of the integrals of the continuous and modulated signal by choosing the threshold such that a single pulse resets the error integrator to zero. 
 	Thus the threshold must be the inverse of the control frequency times the nominal force, i.e., $\epsilon = \frac{f_{nom}}{f_{smpl}}$. 
\end{enumerate} 
Finally, the modulator feeds back the current output value, closing the loop.

\subsubsection{State-Estimation}

This work assumes that the entire state is available for feedback in the previous sections. 
This assumption is not generally satisfied since the \gls{mocap} system only provides pose estimates $(x,y,\theta)$, and numerically differentiating those is not sufficient~\cite{Floris2020}.
Further, all available measurements are subject to noise and thus require some filtering process to improve their quality given some underlying system model. 
The standard for this filter is the \gls{kf}~\cite{Kalman1960}. 
By combining the knowledge of the underlying system and the measurements optimally, in the sense of a quadratic estimation error, the filter smoothes the data and rejects extreme outliers that fall significantly outside the distribution of the specified measurement error. 
This work combines a classical \gls{kf} for the position, its derivative, and the \gls{rw} velocity with a \gls{kf} over the Lie Algebra $SO(2)\times\mathbb{R}$ for the orientation and the angular velocity as given in~\cite{Markovic2017} to achieve full state estimation.

\section{RESULTS}\label{sec:results}

\subsection{Simulation}

Before being evaluated on the physical system, this work evaluates the controller in a Gazebo~\cite{gazebo} simulation that incorporates sensor noise and the slight unevenness of the floor.
The simulated sensors are subjected to \gls{awgn} with variances that match the measured variances of the real system ($\sigma_x^2 = \sigma_y^2 = \SI{1e-5}{\meter^2}$, $\sigma_\theta^2=\SI{1e-5}{\radian^2}$ and $\sigma_{\omega_{RW}}^2 = \SI{1e-4}{(\radian\per\second)^2}$).

The simulation includes a heightmap representing the last openly available measurements from~\cite{ORL} to incorporate the unevenness of the flat-floor, which has a maximal deviation of \SI{1}{\milli\meter} over one meter.
For more details about the simulation, the interested reader is referred to~\cite{bredenbeck2022astra}.

\subsubsection{Stabilization}

\begin{figure}
	\centering
	\includegraphics[width=0.24\textwidth]{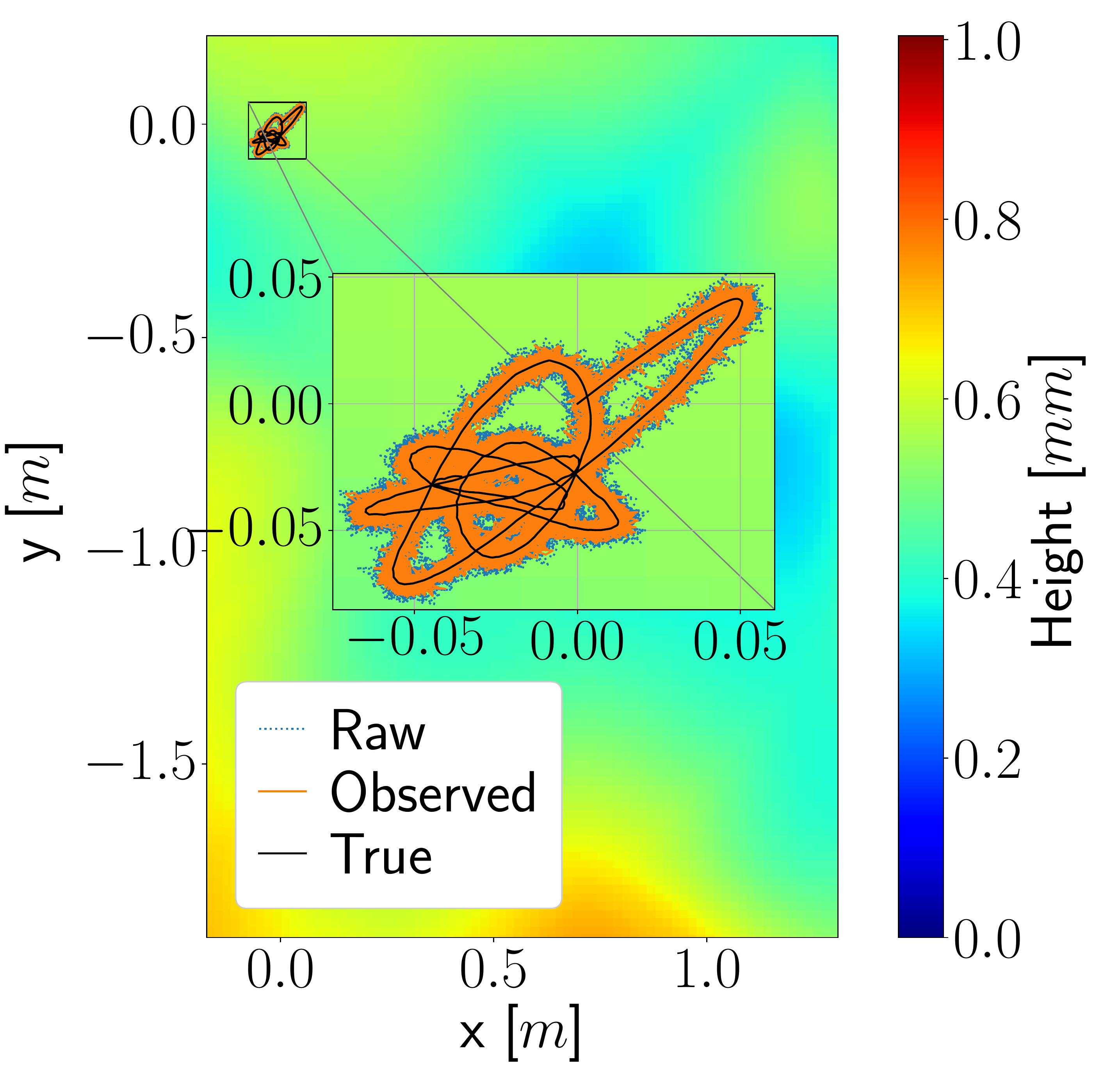}\hfill
	\includegraphics[width=0.24\textwidth]{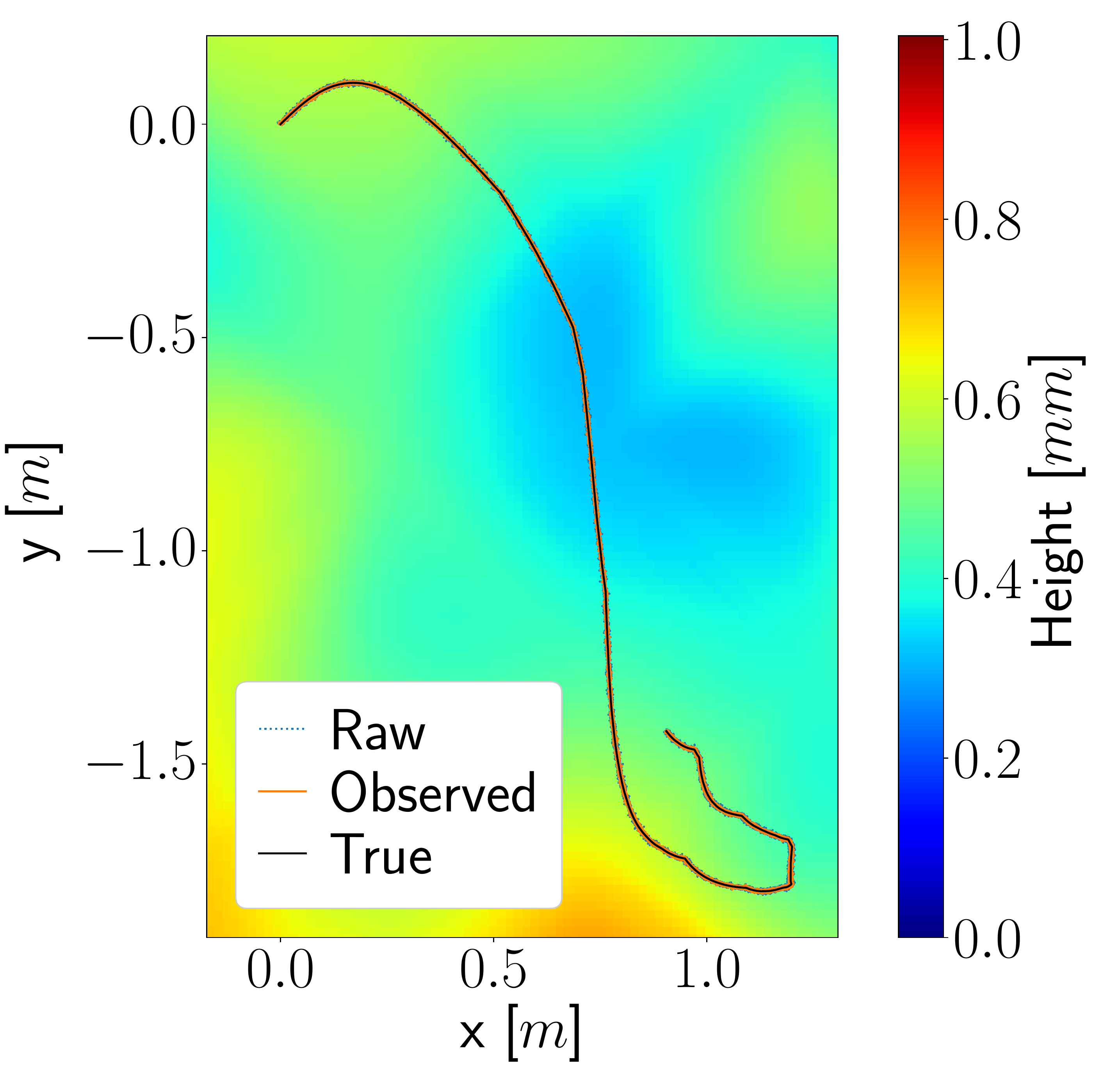}\\
	\includegraphics[width=0.24\textwidth]{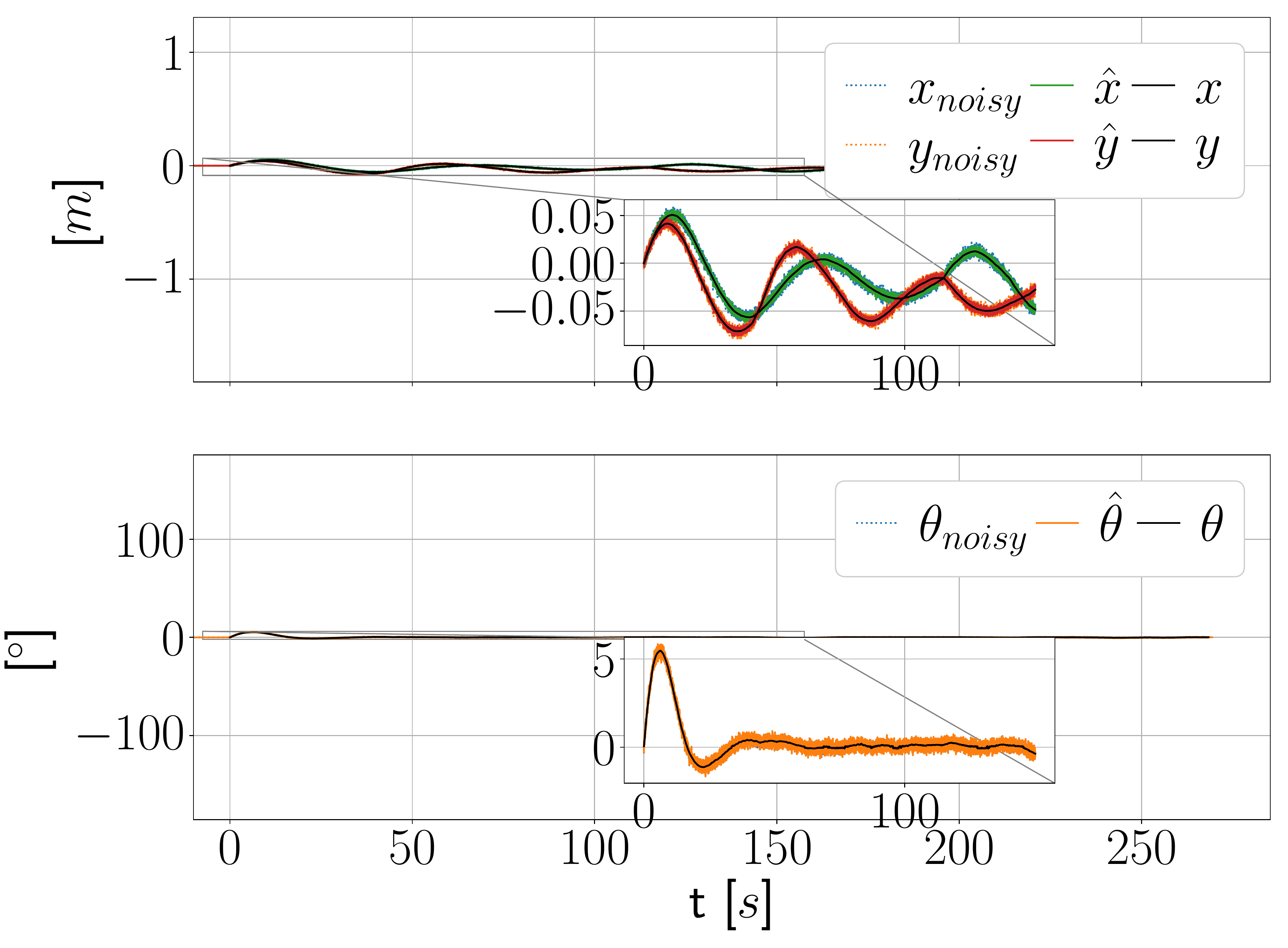}\hfill
	\includegraphics[width=0.24\textwidth]{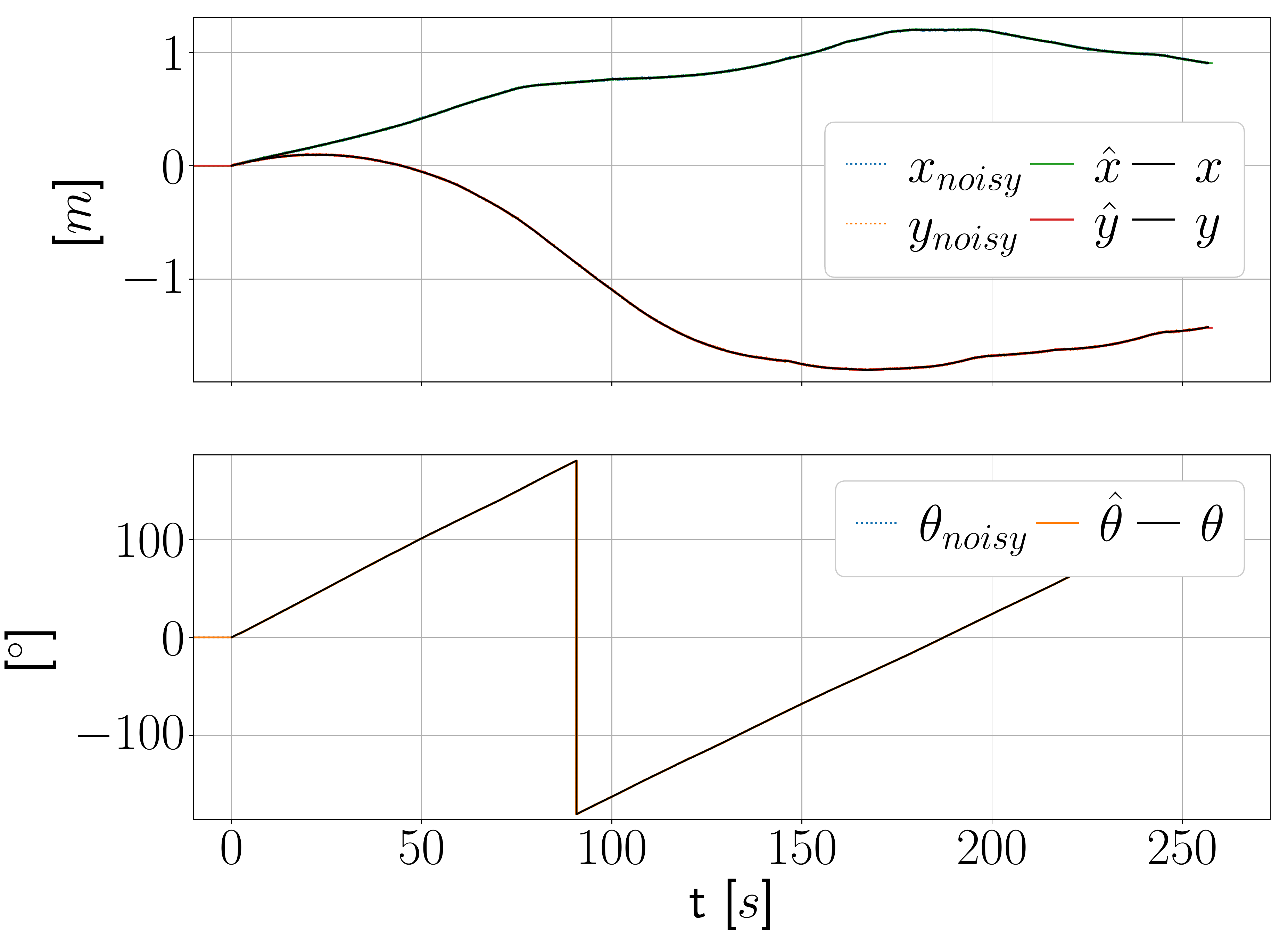}\\
	\includegraphics[width=0.24\textwidth]{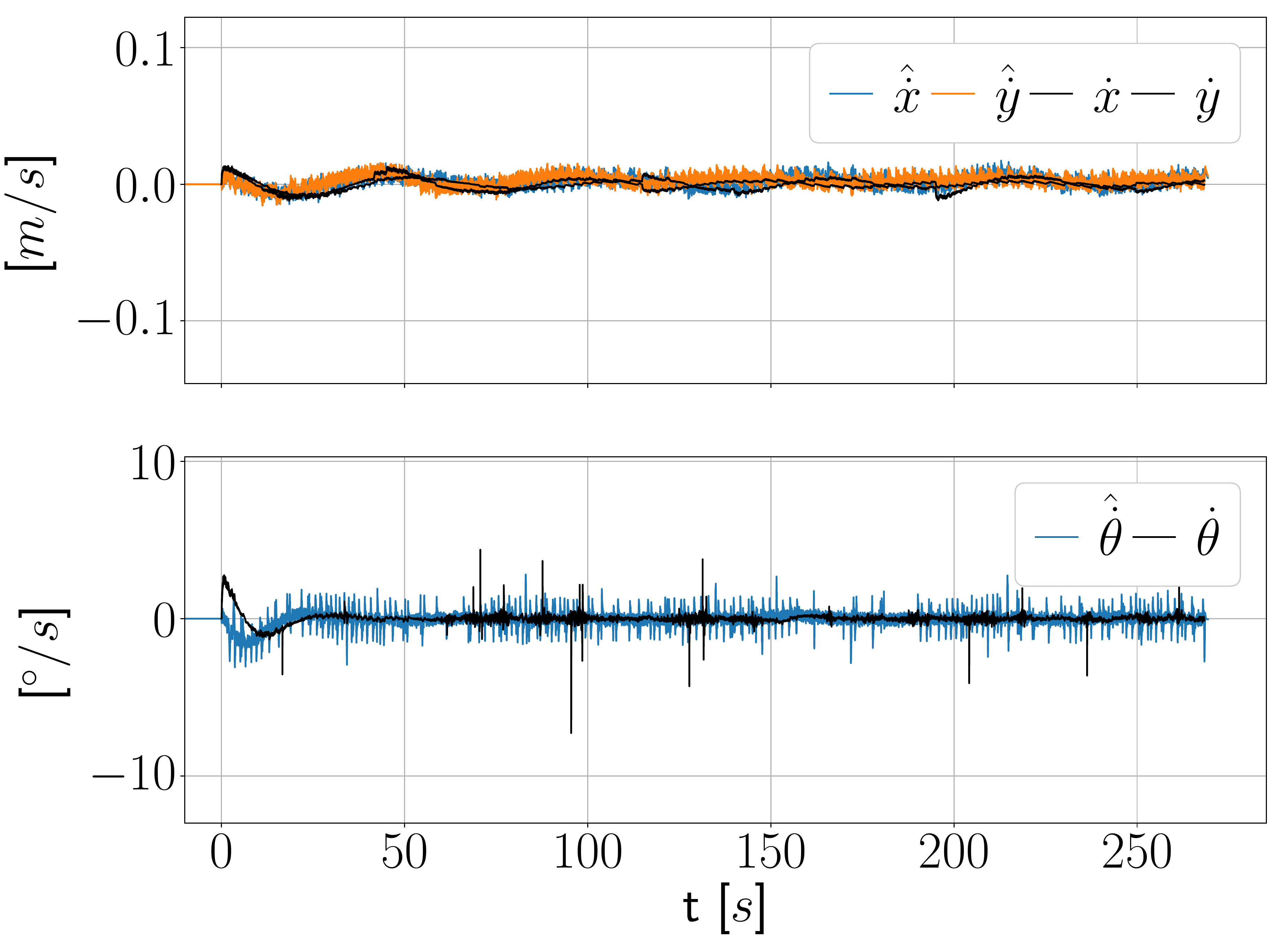}\hfill
	\includegraphics[width=0.24\textwidth]{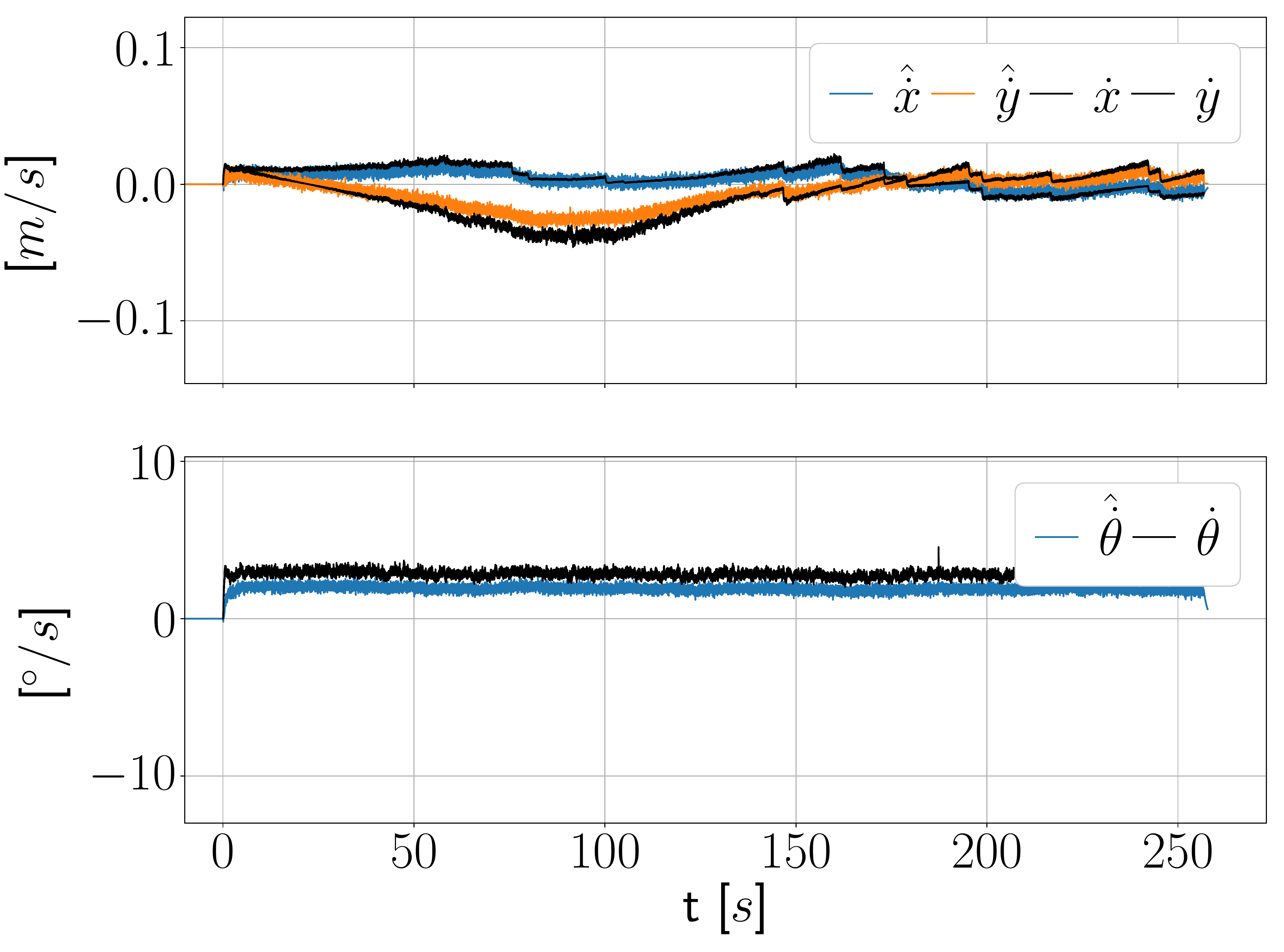}
	\caption{Ground-track, individual coordinates, and velocities of the system responding to a disturbance (at $T=\SI{0}{\second}$) on an uneven floor in simulation. Left: Controller stabilizing the system at the origin. Right: No controller running. An animation of the stabilization process is given at \href{https://youtu.be/KRYcq3VjQUo?t=4}{\url{https://youtu.be/KRYcq3VjQUo?t=4}}.}
	\label{fig:stab-with-response}
\end{figure}

\begin{figure}
	\centering
	\includegraphics[width=0.24\textwidth]{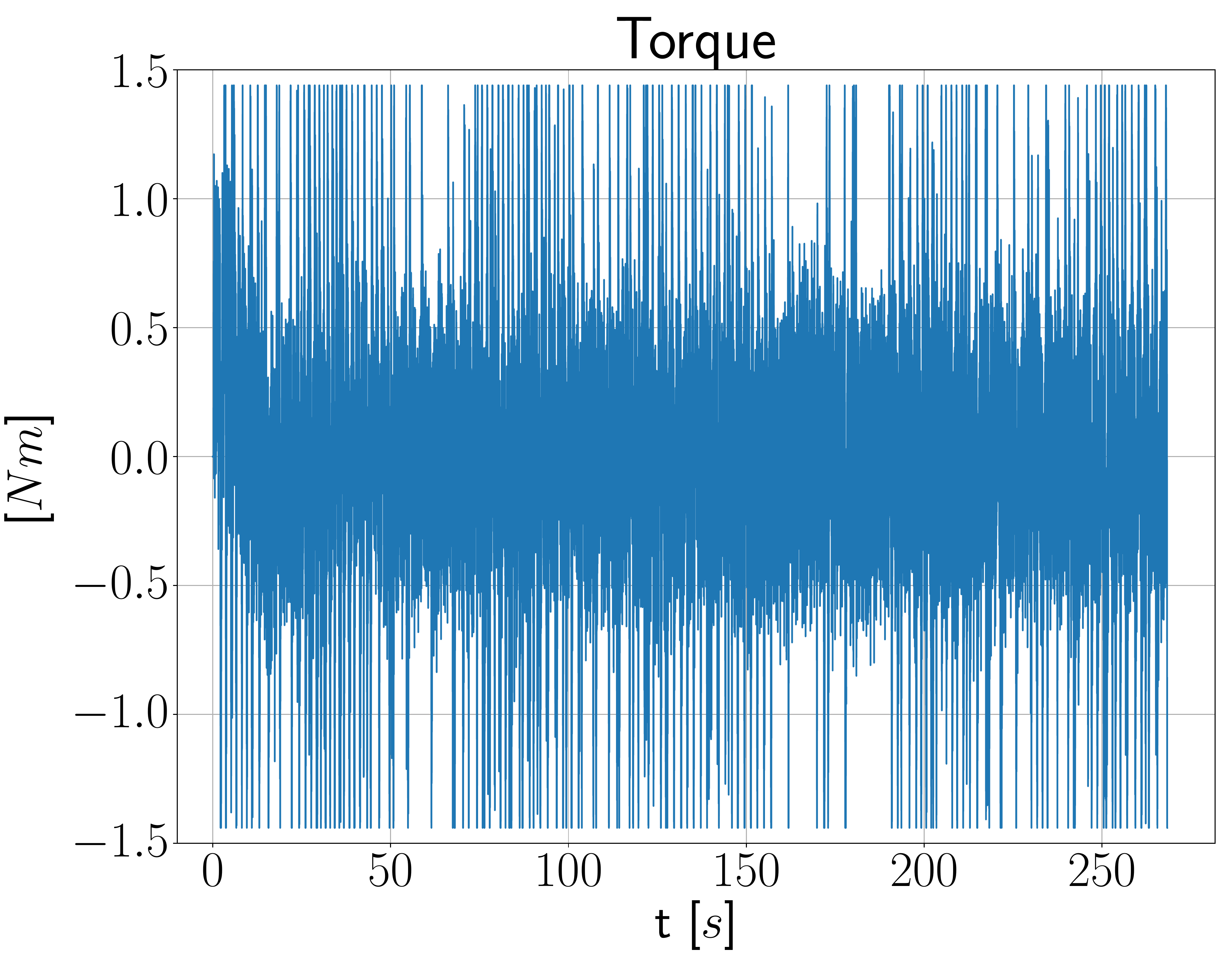}\hfill
	\includegraphics[width=0.24\textwidth]{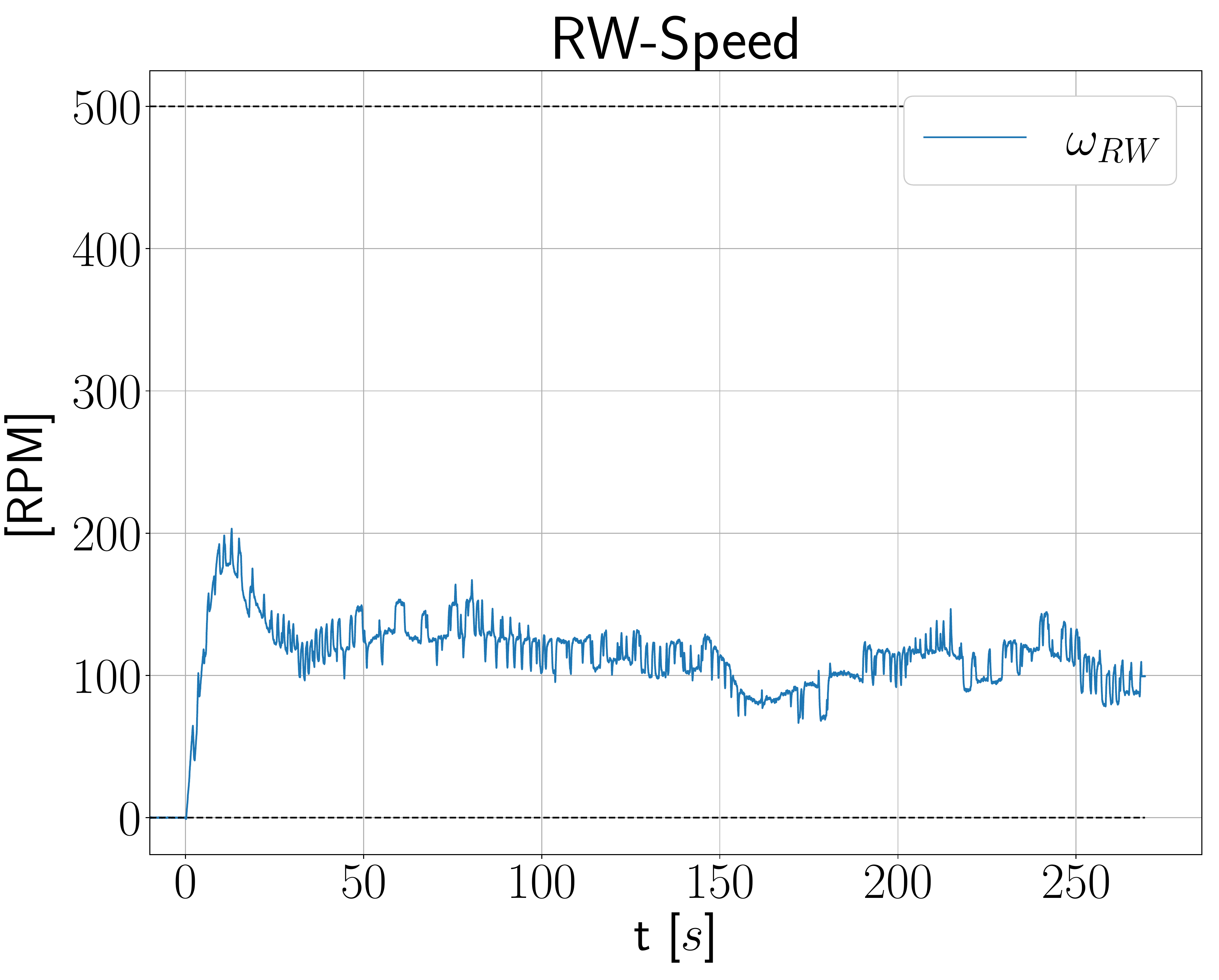}\\
	\includegraphics[width=0.4\textwidth]{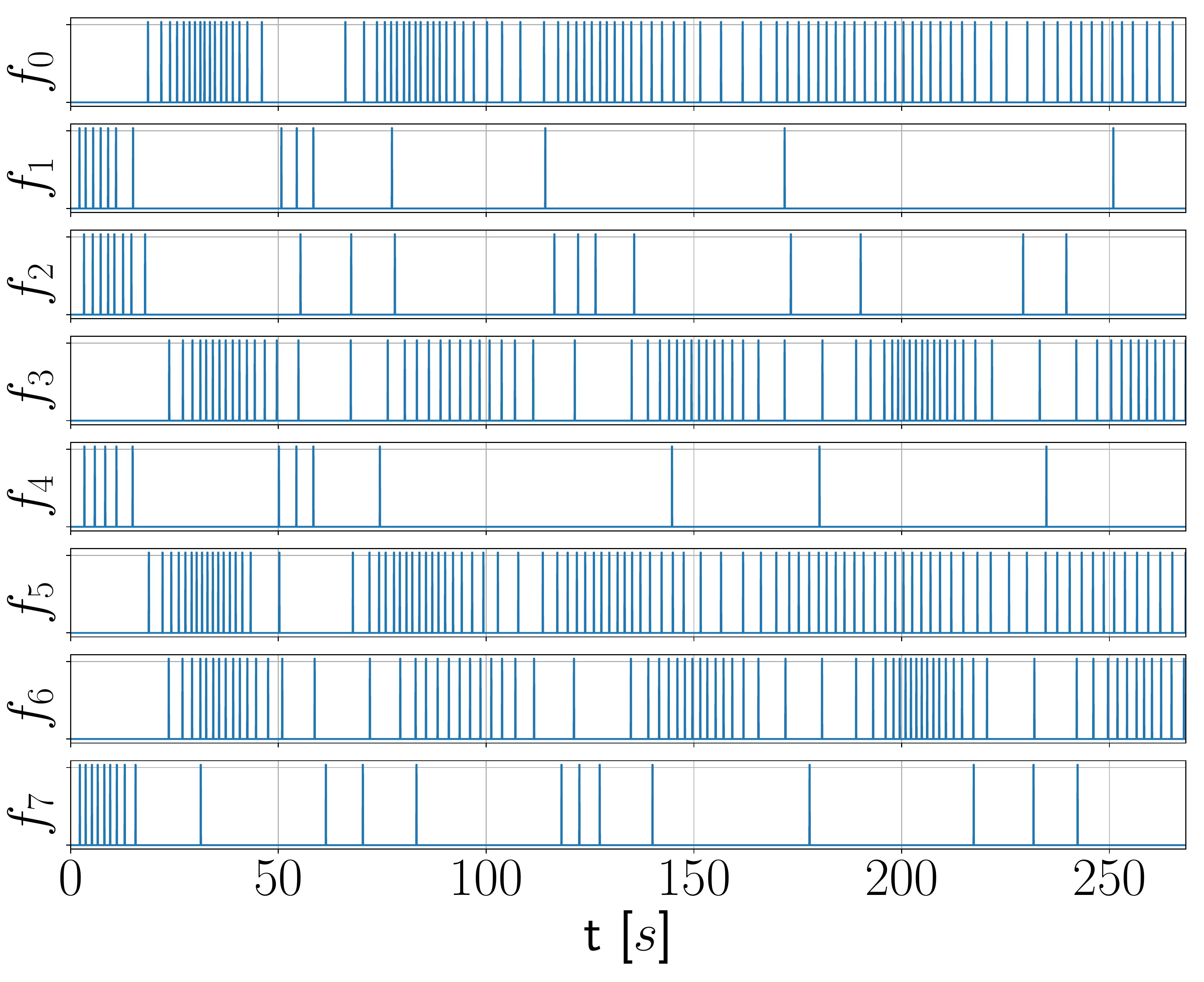}
	\caption{Actuation required to stabilize the system at the origin on an uneven floor in simulation while being subjected to a disturbance (cf. Figure~\ref{fig:stab-with-response}). Top Left: the torque exerted by the motor onto the RW. Top Right: the resulting RW velocity (raw and observed). Bottom: the thruster activity for all eight thrusters.}
	\label{fig:stab-with-actuation}
\end{figure}

The system is simulated while being subjected to the instantaneous disturbance consisting of a force and a torque $\vec{d}= \begin{bmatrix}\vec{f}_d & \vec{\tau}_d\end{bmatrix}$ and lasting for $\Delta t$, where:
\begin{align}
\resizebox{0.42\textwidth}{!}{$
\vec{f}_d = \begin{bmatrix} \SI{5000}{\newton} & \SI{5000}{\newton}\end{bmatrix}^T, \;\; \vec{\tau}_d = \SI{1000}{\newton\meter}, \;\; \Delta t = \SI{0.001}{\second}
$}.
\end{align}  
The response is shown in Figure~\ref{fig:stab-with-response} and the actuation of the controller in Figure~\ref{fig:stab-with-actuation}. 
Accompanying each plot of the controlled system response is a plot of the system response to the disturbance without being controlled to highlight the effect of the controller.

As Figure~\ref{fig:stab-with-response} shows, the controller succeeds in bringing the system back to the origin after the initial disturbance. 
It maintains the system within less than \SI{10}{\centi\meter} and \SI{6}{\degree} of the origin. 
The unactuated system, on the other hand, drifts over \SI{1.5}{\meter} and performs more than a full rotation. 
The offset from the origin of the controlled system is attributed to two factors: the thruster's binary nature and the unevenness of the floor. 
The continuous actuation needs to accumulate in the modulator until a firing is triggered, propelling the system through the origin, and the process repeats in the opposite direction. 
This well-known behavior for thruster-based systems is referred to as a limit cycle~\cite{Jeon2012}.
At the same time, the controller needs to compensate for the disturbances induced by the flat-floor, further adding to the oscillations about the origin.
In Figure~\ref{fig:stab-with-actuation} the torque clearly shows ``bang-bang'' behavior. 
Since the inertia of the \gls{rw}s is two orders of magnitude smaller than the system's inertia, the controller demands high torques quickly alternating between maximal and minimal torque.  
However, the \gls{rw} velocity remains within its allowable bounds. 
Also, in Figure~\ref{fig:stab-with-actuation}, the thrusters 1, 2, 4, and 7 initially show high activity since they are the ones pointed in negative $x$ and $y$ direction, thus counteracting the disturbance.
Afterward, the thrusters show alternating behavior corresponding to the small oscillations about the origin. 
Finally, the thrusters regularly fire to move the system towards the positive $y$ direction (Thrusters 0 and 5) and slightly less frequently towards the positive $x$ direction (Thrusters 3 and 6), which corresponds to the gradient of the floor at the origin. 

\subsubsection{Monte Carlo}\label{subsec:monte-carlo}

\begin{figure}
	\centering
	\vspace{0.15cm}
	\includegraphics[trim=12cm 0cm 2cm 0cm, clip=true, height=3.8cm]{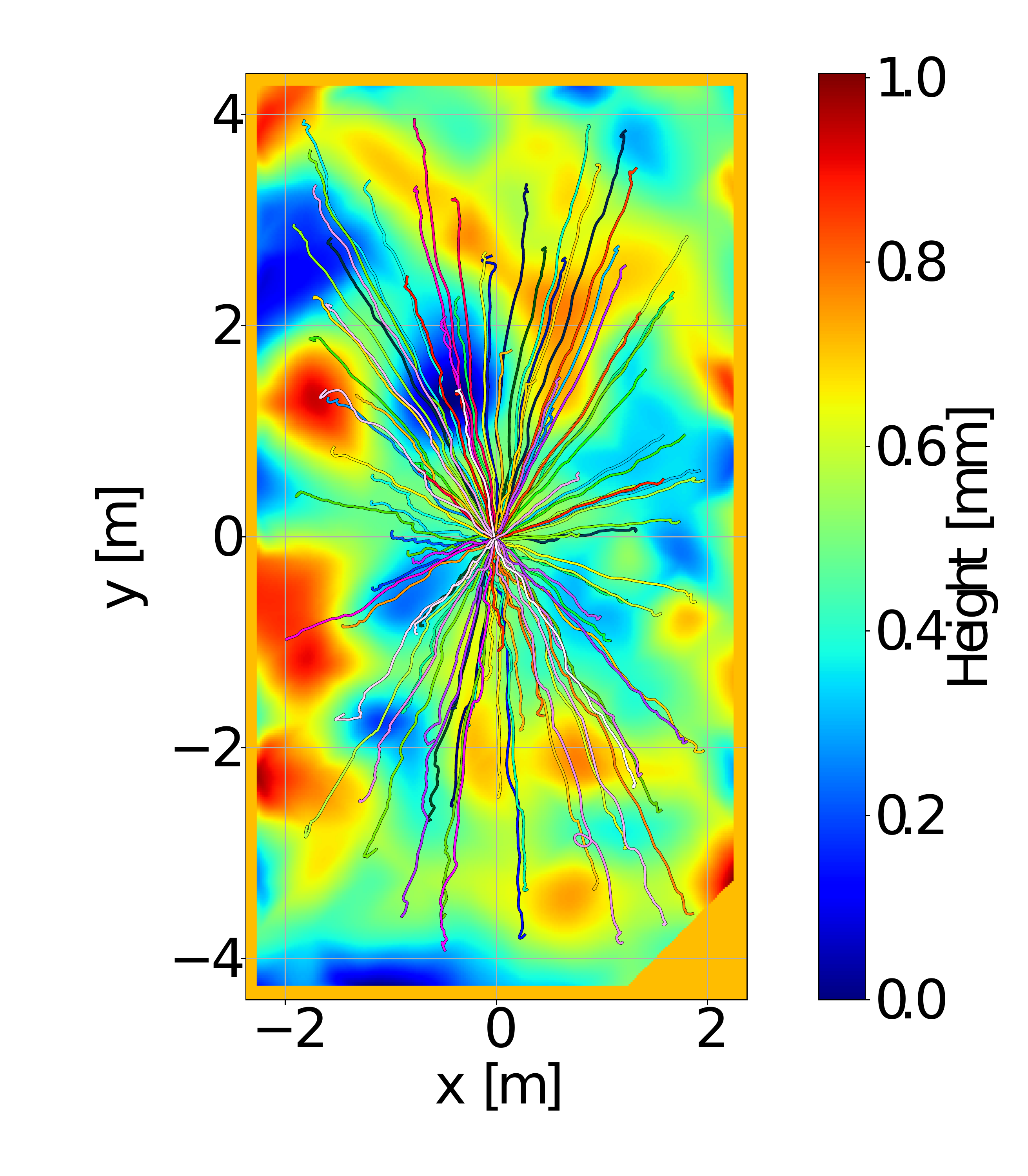}
	\includegraphics[height=3.8cm]{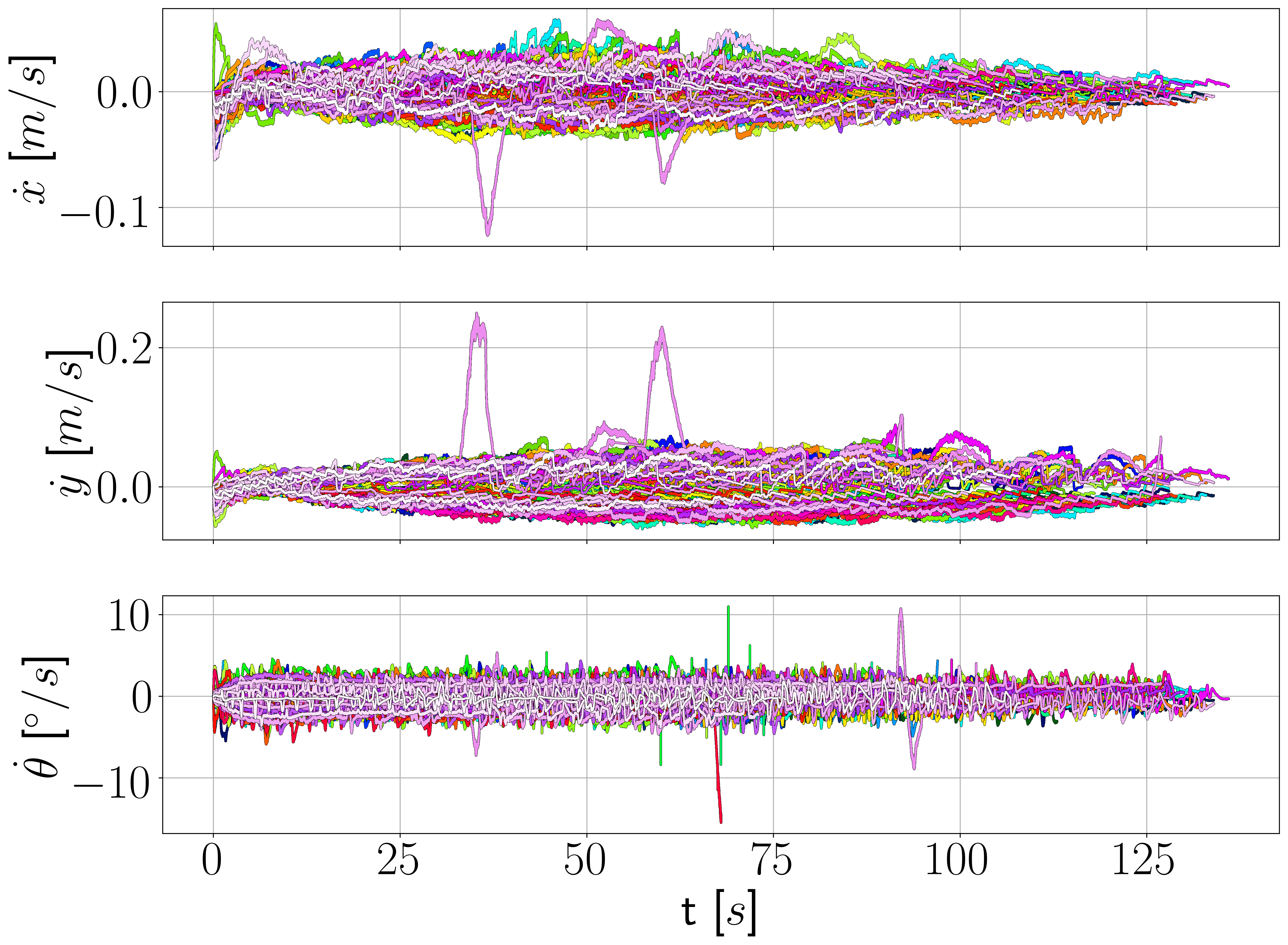}\\
	\includegraphics[trim=0cm 1cm 0cm 0cm, clip=true, width=0.45\textwidth]{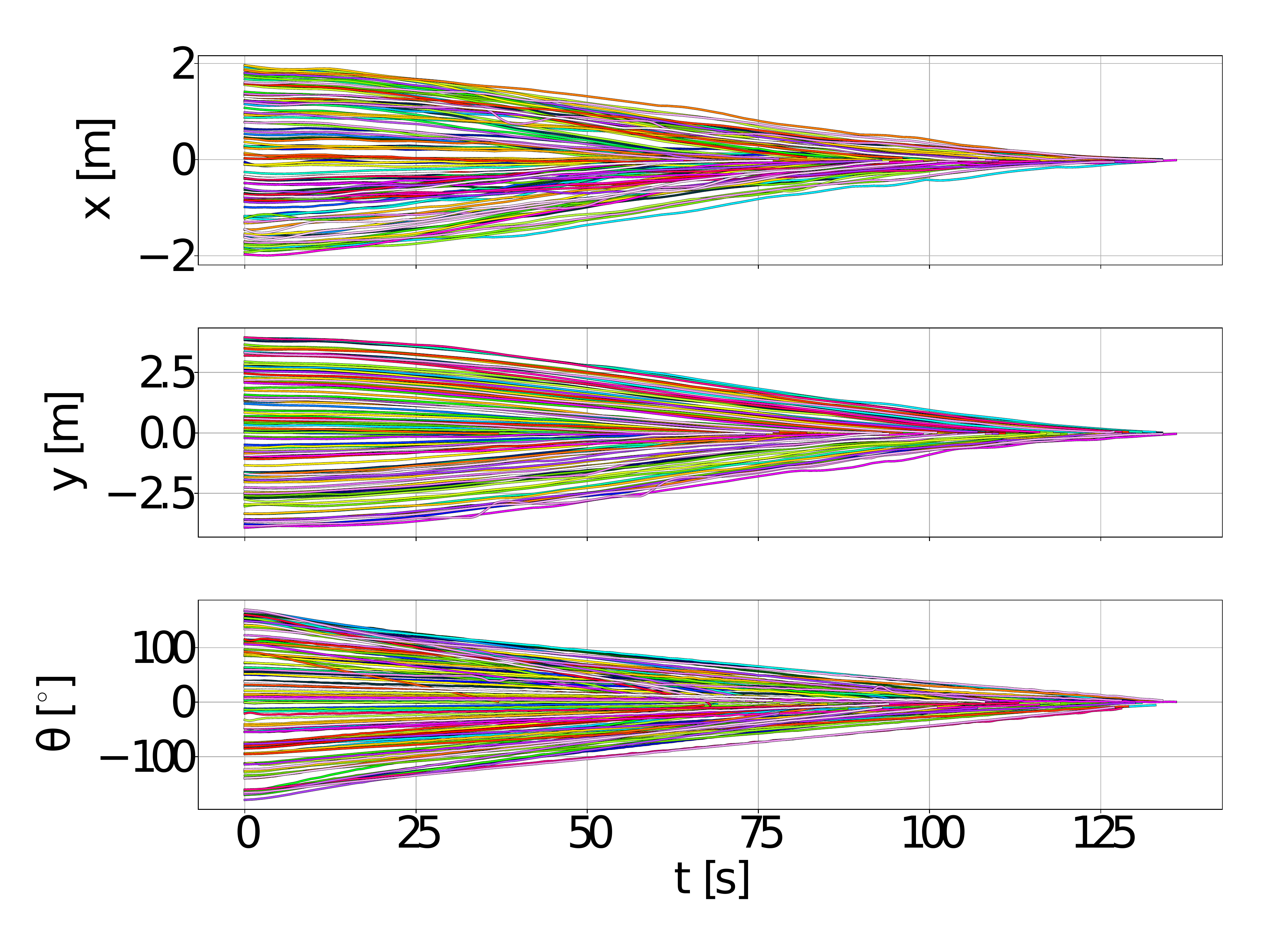}
	\caption{Monte Carlo simulation of following trajectories to the origin. Each optimal trajectory is computed using $N=100$ knot-points.}
	\label{fig:monte-carlo} 
\end{figure}

To demonstrate the generalizability of the controller to different initial locations on the flat-floor the system also undergoes a Monte Carlo test simulation.
For a large ($n = 100$) number of episodes, the robot spawns at random initial poses, whereby $x, y$, and $\theta$ result from drawing from uniform random distributions that span the range $[-2, 2]$, $[-4, 4]$, and $[-\pi,\pi]$ respectively. 
The controller computes the optimal trajectory to the origin and starts following it.
An episode is successful when the euclidean distance to the origin, the euclidean velocity as well as the angular error and velocity are smaller than the threshold $\vec{\epsilon}$:
\begin{align}
	&\begin{bmatrix}
		\epsilon_{lin}&\epsilon_{ang}&\epsilon_{lin, vel} &\epsilon_{ang, vel}
	\end{bmatrix} \nonumber\\
	&=\begin{bmatrix}
		\SI{0.05}{\meter}&\SI{0.05}{\meter\per\second}&\SI{0.05}{\radian}&\SI{0.05}{\radian\per\second}
	\end{bmatrix}
\end{align}
Figure~\ref{fig:monte-carlo} shows the results of the Monte Carlo test. 
During the Monte Carlo simulation, the controller commands the system to the origin from all 100 tested initial conditions. 
As seen in Figure~\ref{fig:monte-carlo} the trajectories are mostly straight lines with some minor deviations caused by the uneven ground, combined with the noise of the system and the binary thrusters.
One particular trajectory (purple, bottom right of the ground-track) overshoots a desired state along the trajectory at approximately \SI{38}{\second}. 
Since the controller only attempts to control the system to the desired state at some time (single planning approach), it returns to a previous location resulting in a loop. 
This loop also corresponds to the spikes in velocities (purple).
All trajectories reach the desired region at the origin and slow down to the desired maximal velocity in less than \SI{140}{\second}.

\subsection{Physical System}

\subsubsection{Experimental Setup}

The \gls{rw} experiences stiction at low revolutions per minute.
Thus, to avoid issues, the \gls{rw} is spun up to half its rated velocity before each trajectory.
From this state, the experimenter places the system manually at the origin of the coordinate system before starting the trajectory follower. 

\subsubsection{Results}  

\paragraph{Stabilization}

\begin{figure}
	\centering
	\includegraphics[height=3.5cm]{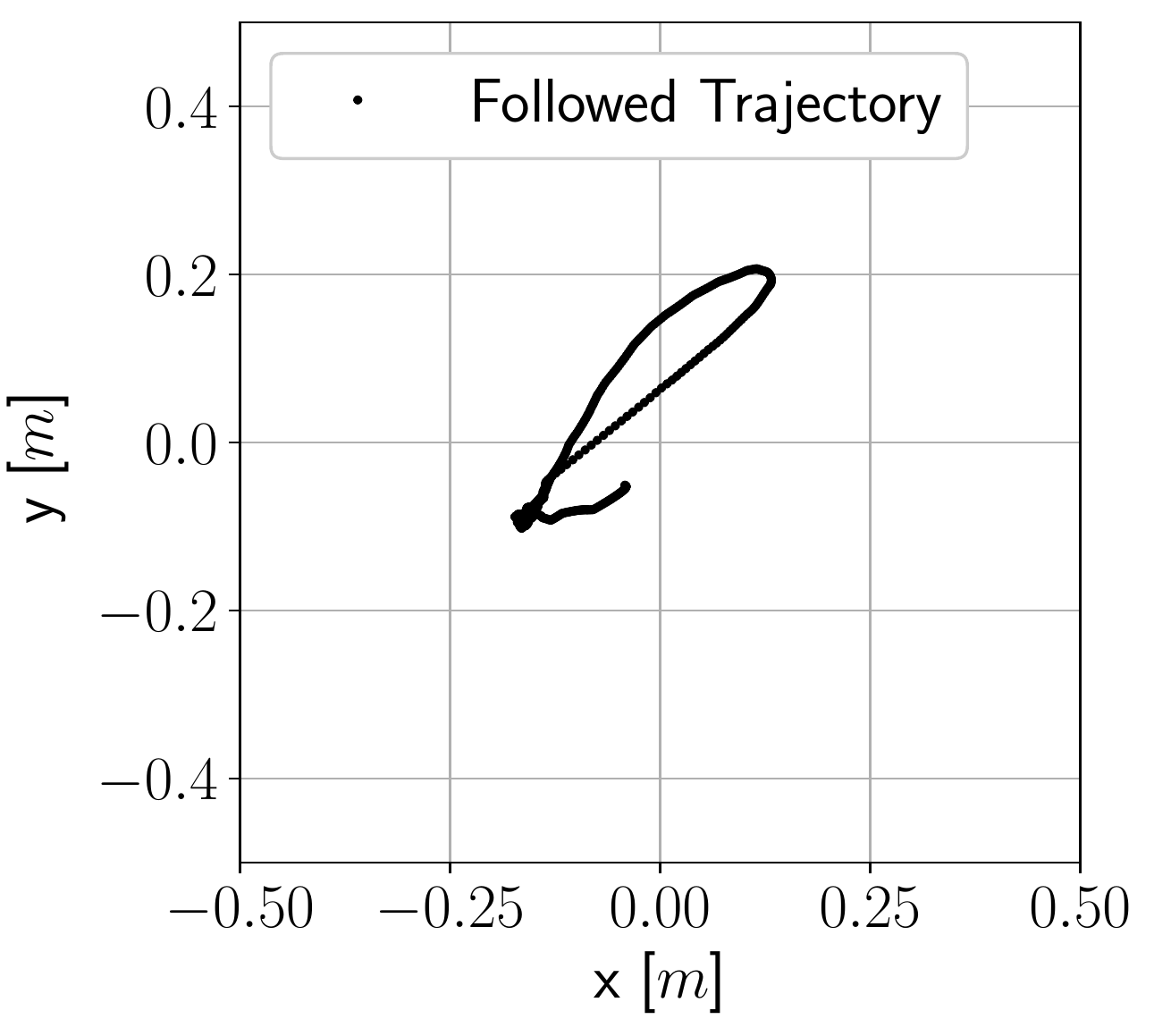}\hfill
	\includegraphics[height=3.5cm]{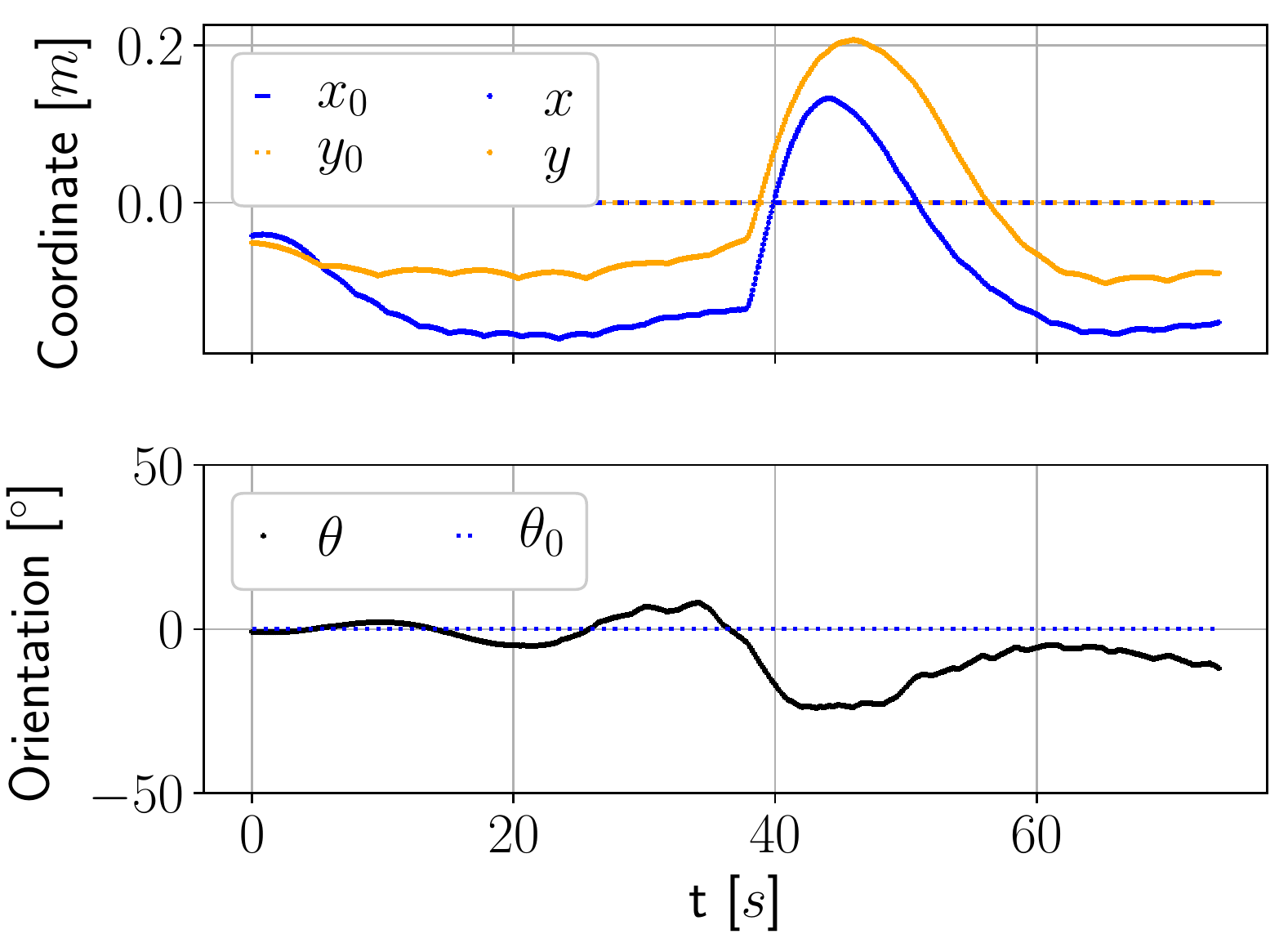}
	\caption{Ground-track and individual coordinates of the controller stabilizing the physical system at the origin. 
	On the right the desired value is indicated as a dashed line.
	A video is given at \href{https://youtu.be/KRYcq3VjQUo?t=168}{\url{https://youtu.be/KRYcq3VjQUo?t=168}}.}
	\label{fig:real-stabilize-gt}
\end{figure}

\begin{figure}
	\centering
	\vspace{0.15cm}
	\includegraphics[width=0.24\textwidth]{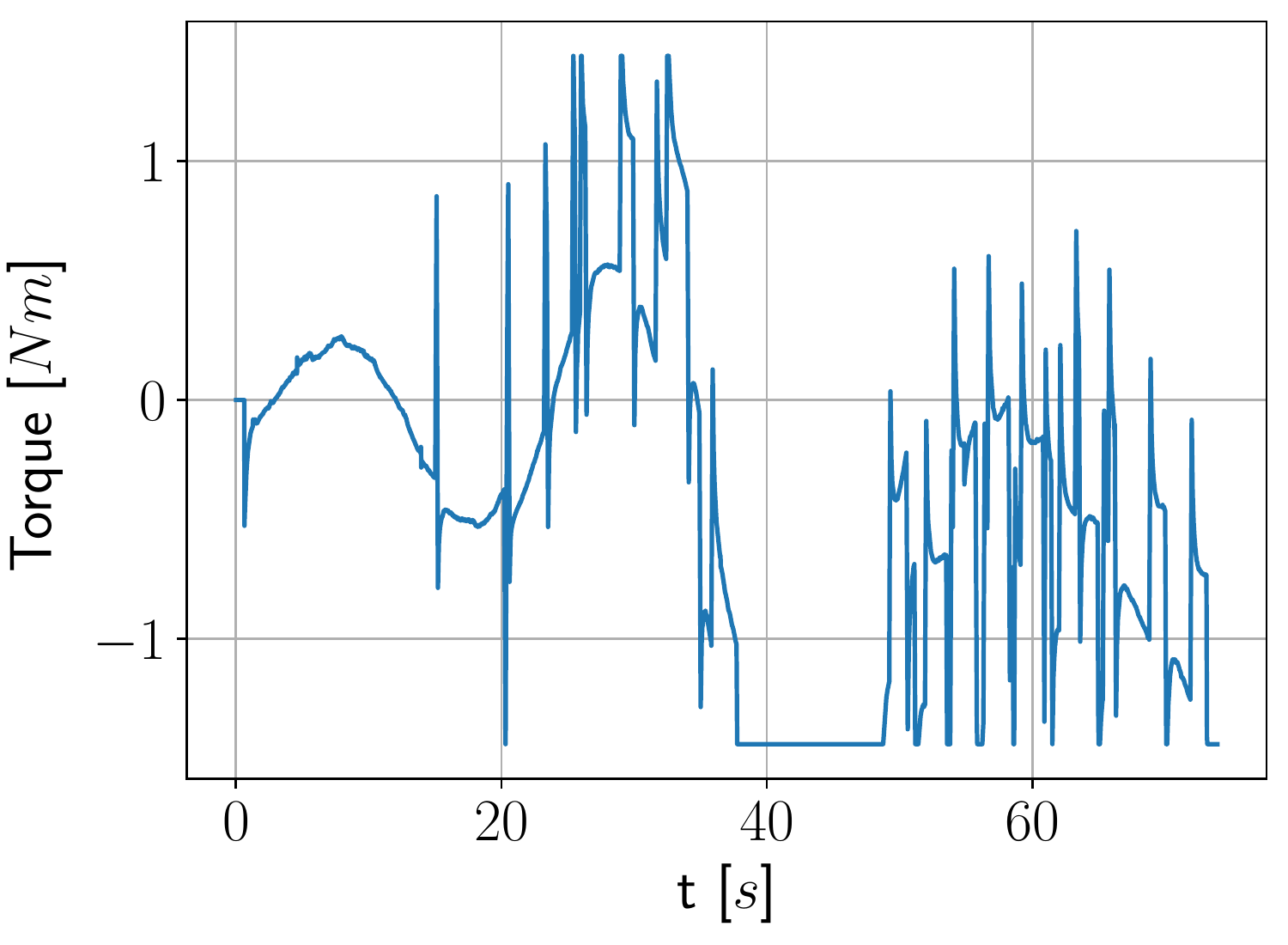}\hfill
	\includegraphics[width=0.24\textwidth]{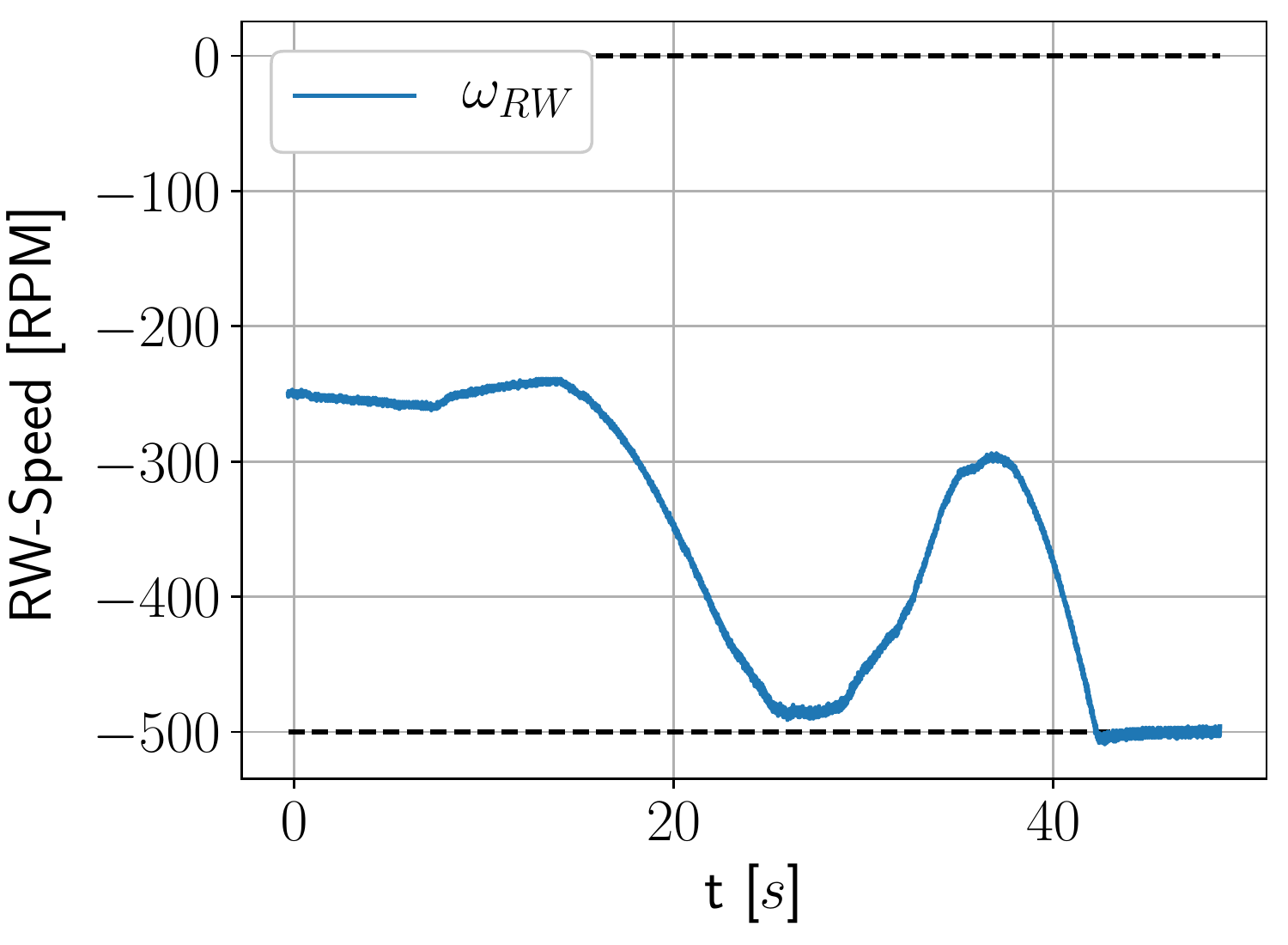}\\
	\includegraphics[width=0.3\textwidth]{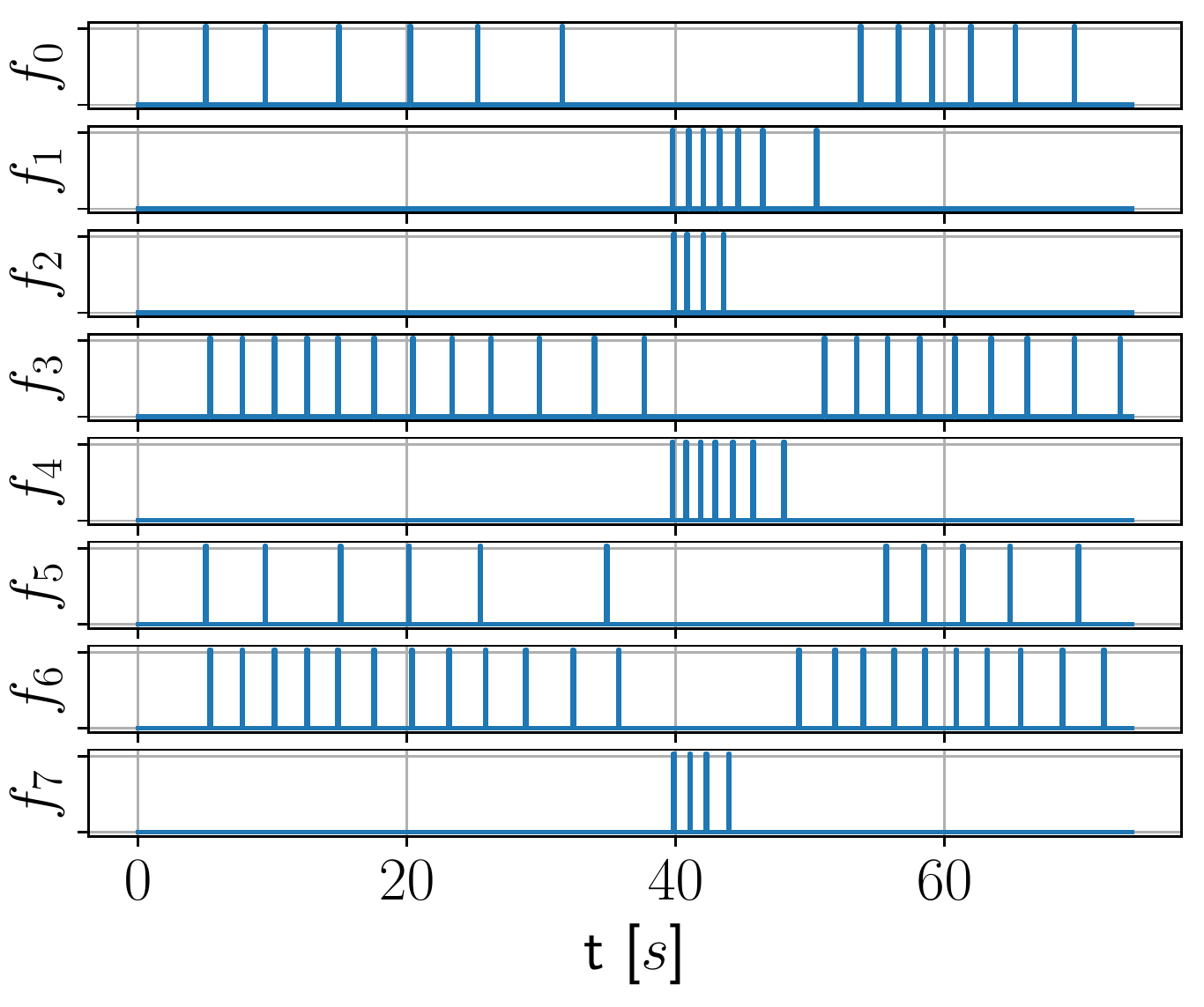}
	\caption{Actuation of stabilizing the physical system (cf. Figure~\ref{fig:real-stabilize-gt}).}
	\label{fig:real-stabilize}
\end{figure}

This first section also evaluates the system's stabilization capabilities for the physical system. 
The disturbance is added by manually pushing the system.
The results are shown in Figures~\ref{fig:real-stabilize-gt} and~\ref{fig:real-stabilize}.


The system generally exhibits the tendency to drift in negative $x$ and $y$ directions in all experiments.
This tendency implies a slope pointing in this direction, which the heightmap from~\cite{ORL} further supports.
During the stabilization process, the system initially stabilizes within \SI{15}{\centi\meter} and \SI{15}{\degree} of the origin before the disturbance. 
The disturbance moves the system about \SI{35}{\centi\meter}, and within \SI{30}{\second} the system stabilizes in the initial region again.
Figure~\ref{fig:real-stabilize} shows that very little thrusting is required to achieve this result, firing no thruster more than once every two seconds during the entire process. 
The \gls{rw}, on the other hand, saturates rather quickly after the disturbance. 
It quickly reaches its maximum rotational velocity as it attempts to absorb the entire angular momentum put into the system by the disturbance.
Afterward, the thrusters perform the attitude control themselves; thus, the pointing accuracy degrades as larger errors are required to trigger a pulse since thruster usage is penalized more than \gls{rw} acceleration.

\paragraph{Straight-Line Trajectory}

\begin{figure}
	\centering
	\vspace{0.1cm}
	\includegraphics[height=4.025cm]{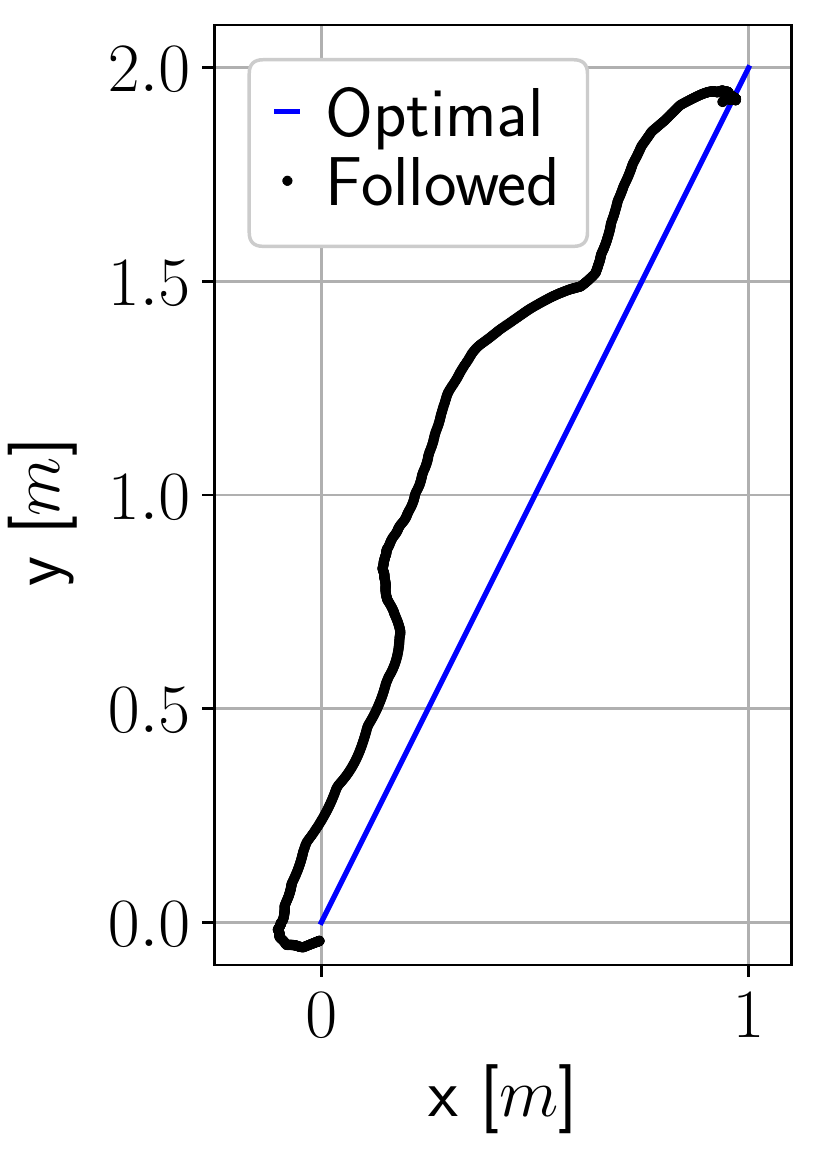}\hfill
	\includegraphics[height=4.025cm]{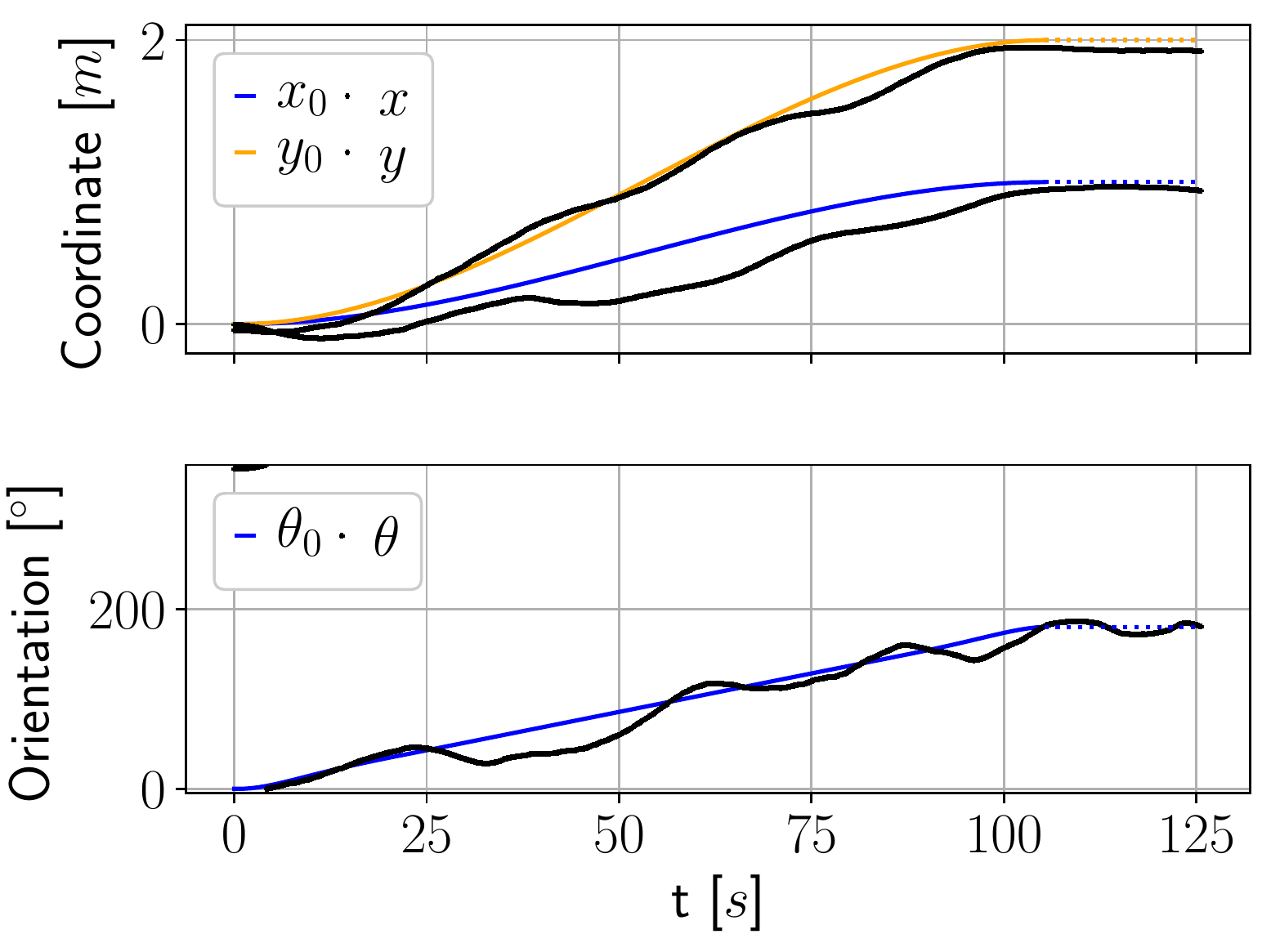}
	\caption{Ground-track and individual coordinates of the controller following a straight-line trajectory on the physical system. 
	After reaching the final pose of the trajectory the desired value is indicated as a dashed line.
	A video is given at \href{https://youtu.be/KRYcq3VjQUo?t=193}{\url{https://youtu.be/KRYcq3VjQUo?t=193}}.}
	\label{fig:real-straight_line-gt}
\end{figure}

\begin{figure}
	\centering
	\vspace{0.1cm}
	\includegraphics[width=0.24\textwidth]{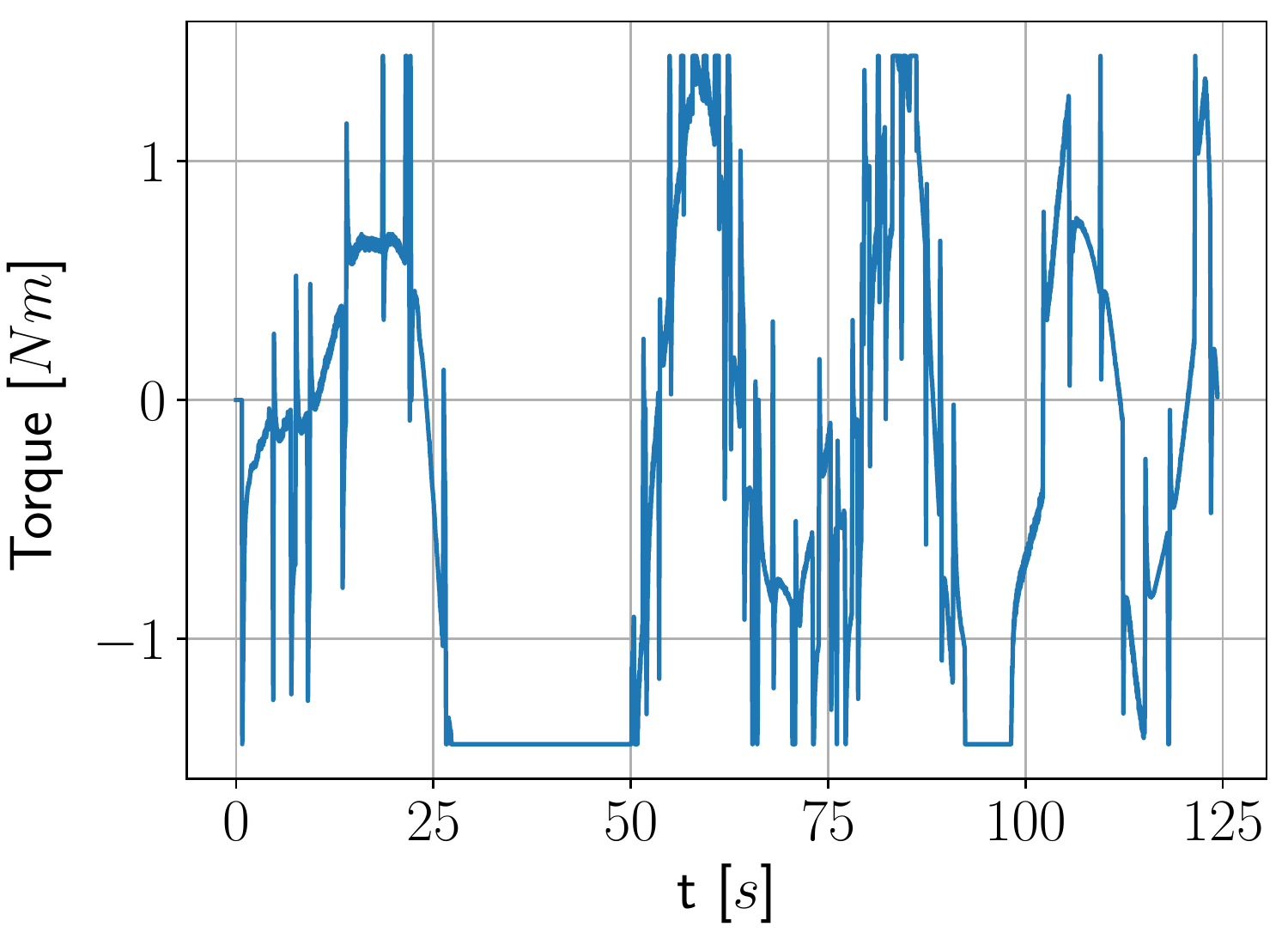}\hfill
	\includegraphics[width=0.24\textwidth]{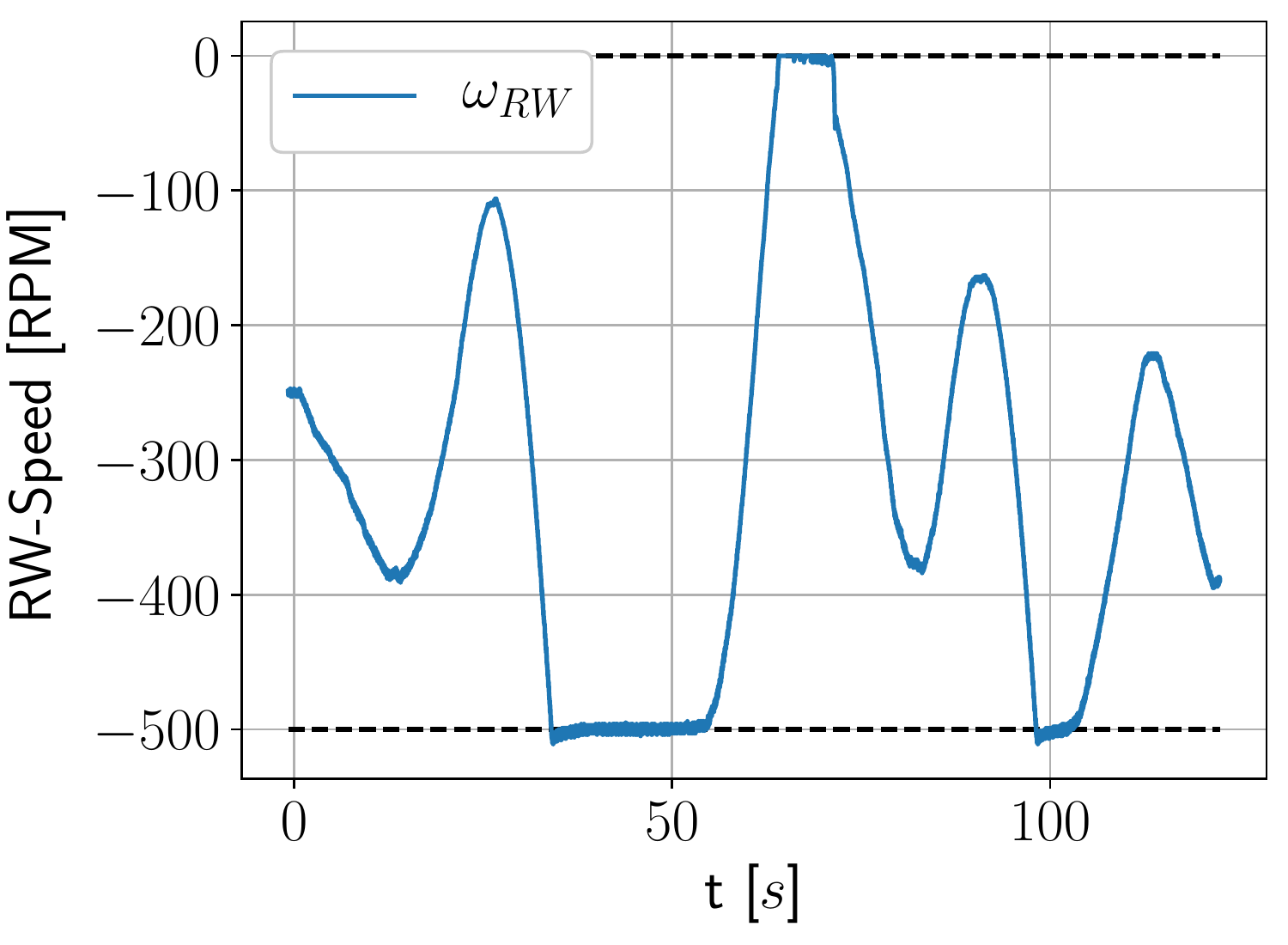}\\
	\includegraphics[width=0.3\textwidth]{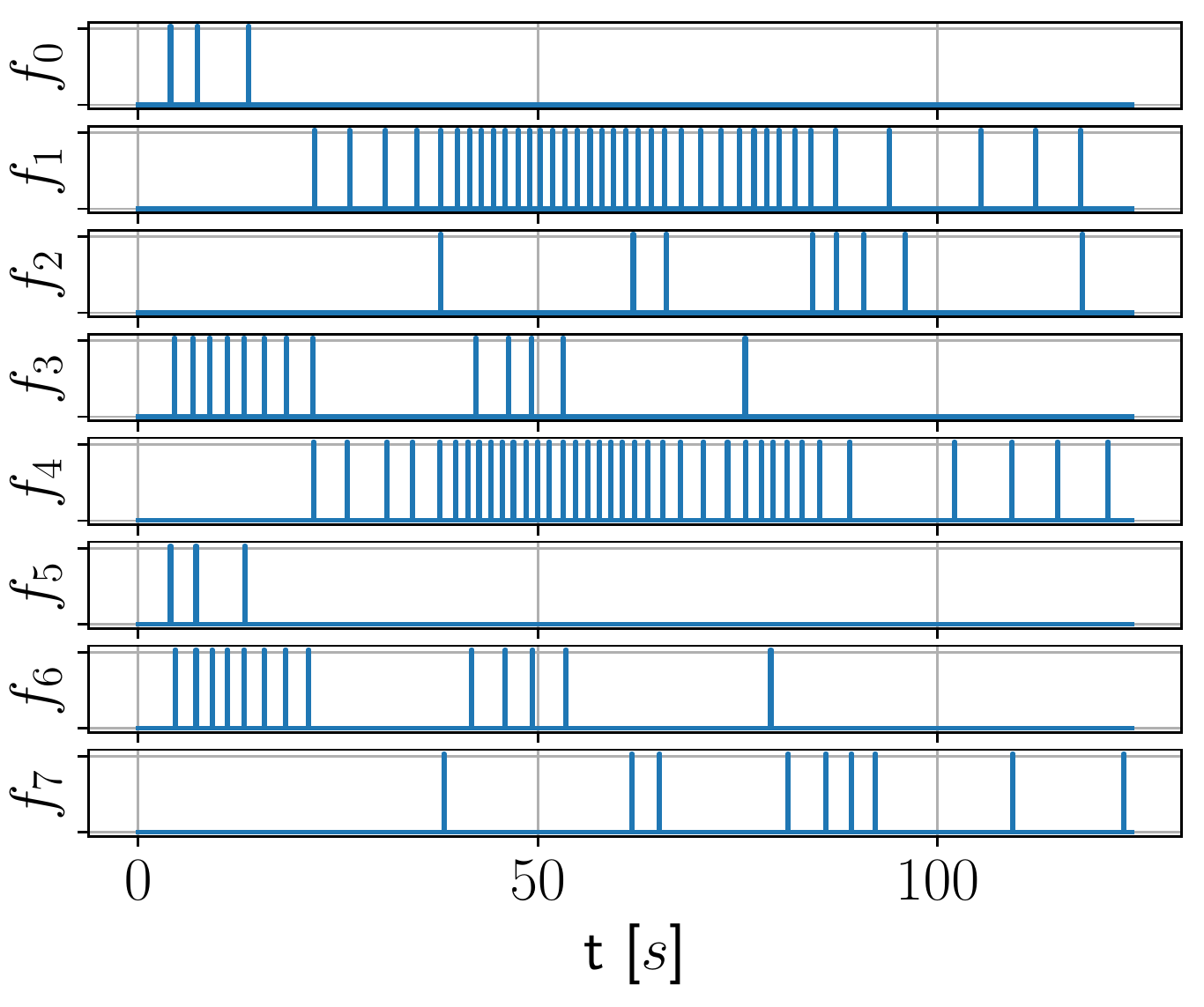}
	\caption{Actuation of following the straight-line trajectory (cf. Figure~\ref{fig:real-straight_line-gt}) on the physical system.}
	\label{fig:real-straight_line}
\end{figure}

Further, this section evaluates the physical system's performance following a straight-line trajectory (similar to the ones from section~\ref{subsec:monte-carlo}). 
The results are shown in Figures~\ref{fig:real-straight_line-gt} and~\ref{fig:real-straight_line}.

The average Euclidean and angular error of the physical system to the desired trajectory are \SI{0.325}{\meter} and \SI{23.8}{\degree} respectively, in a trajectory covering \SI{2.2}{\meter} and \SI{180}{\degree}.
In particular, the error in the $x$-axis is significant. 
This error is due to the slope in that region of the flat-floor, with a gradient in the negative $x$-direction.
As in the stabilization case, the heightmap shows a local minimum adjacent to the trajectory.

Further, the \gls{rw} saturates at both ends of the allowable range throughout the trajectory. 
At each instance, the orientation deviates more significantly from the desired orientation, implying that large desired changes in angular momentum of the entire system quickly lead to \gls{rw} saturation.
The controller then performs orientation control using only thrusters, thus, losing precision.
However, at the end of the trajectory, the \gls{rw} recovers from saturation, and the pointing accuracy at the final state is within \SI{10}{\degree}.
The thrusters were firing appropriately, never exceeding one fire per second for an individual thruster.
Moreover, when considering the thruster on-times, as depicted in Table~\ref{tab:thruster-on-time}, it shows an increase of less than one order of magnitude for the overall thruster on-times compared to the optimal value.
Some individual thrusters (0, 5) even decrease on-time, whereas others increase by one or multiple orders of magnitude while compensating for uneven ground.
Note that the long on-times for thrusters 1 and 4 stem from compensating the previously mentioned error in $x$-direction since these thrusters are enacting a force in positive $x$-direction throughout the middle section of the trajectory.

\begin{table}
 	\centering
	\vspace{0.0cm}
 	\caption{Optimal and real thruster on-time (cf. Figure~\ref{fig:real-straight_line-gt}).}
 	\resizebox{0.48\textwidth}{!}{
 	\begin{tabular}{@{}c|cccccccc|l@{}}\hline
 	Thruster & 0 & 1 & 2 & 3 & 4 & 5 & 6 & 7 & $\Sigma$\\\hline\hline
 	Optimal & \SI{0.499}{\second} & \SI{0.004}{\second} & \SI{0.061}{\second} & \SI{0.296}{\second} & \SI{0.004}{\second} & \SI{0.499}{\second} & \SI{0.296}{\second} & \SI{0.061}{\second} & \SI{1.72}{\second}\\
 	Real & \SI{0.30}{\second} & \SI{3.40}{\second} & \SI{0.70}{\second} & \SI{1.30}{\second} & \SI{3.40}{\second} & \SI{0.30}{\second} & \SI{1.30}{\second} & \SI{0.70}{\second} & \SI{11.40}{\second}\\\hline
 	\end{tabular}
 	}
 	\label{tab:thruster-on-time}
\end{table} 

\paragraph{Discussion}

The experiments demonstrate that the general behavior of the simulation and the physical system is similar, indicating that the simulation is a realistic representation of the scenario.
The controller manages to stabilize the system and follow a pre-computed trajectory.
However, the deviations from the set values are more significant in the physical system.
Besides shortcomings in the simulation, such as the idealized thrusters, the limited heightmap resolution, and a single ground-contact point, other factors that influence this difference stem from the hardware.
The most significant factors that influence the overall system performance are the lack of precise system identification and limited control authority.
The former is outside the scope of this work; thus, it relies on previous inertial measurements of the system.
An offset here contributes to the size of the limit-cycle the system exhibits around the desired orientation. 
This effect is particularly powerful since the control architecture combines the controller with a \gls{kf}, such that the model errors accumulate and deteriorate performance~\cite{doyle1978}.
Improving this model regarding inertial parameters, thrust, and thrust vectors for individual thrusters will significantly improve the system performance.

The latter is a consequence of the initial design of the floating platform.  
The system's high mass means that the thrusters can hardly compensate for the effects of any unevenness of the floor. 
For example, on a slope of \SI[per-mode=fraction]{1}{\milli\meter\per\meter} the system experiences a constant disturbance force of approximately \SI{2.2}{\newton}. 
In an ideal scenario aligning two thrusters exactly with this disturbance force, they can provide approximately \SI{20}{\newton} in the opposing direction. 
Thus to compensate for the disturbance force, the thrusters would have to have an on-time of at least 10\%.
This firing rate is already larger than the firing rate observed for the followed straight-line trajectory.

Additionally the \gls{rw} saturates very fast.
Given the inertias of the entire system and the \gls{rw} it can compensate only for a slight change in angular velocity of approximately \SI{6}{\degree\per\second}. 
Despite the trajectory planner incorporating this limitation, any disturbance that imposes an equivalent torque on the system quickly saturates the \gls{rw} and thus degrades the control authority of the system.
One source of such disturbance is the uneven floor. 
Another is an unequal nominal force of individual thrusters. 
When firing two thrusters that point the same way (e.g., thrusters 0 and 5), the exerted wrench should only contain a force pointing in one direction. 
However, assuming a difference between the two thrusters, it also contains a disturbance torque $\tau_d$. 
If this difference is only 10\% and one assumes the 10\% thruster on-time to compensate for the floor unevenness, this will saturate the \gls{rw} within \SI{34}{\second} of operation.

All in all, the evaluation shows that this control architecture, while limited by immaturity aspects, is effective for a free-floating system with binary actuation constraints.

\section{CONCLUSION \& SUMMARY}\label{sec:conclusion}

Three \gls{dof} floating platforms are a good way of partially emulating microgravity environments as they are present in space applications.
This work introduces a controller for one of the heaviest floating platforms in Europe located within the \gls{orl} at ESTEC, \gls{esa}. 

First, this work proposes a dynamic model of the overall system, actuated by eight solenoid-valve thrusters and one \gls{rw}.
Said controller consists of two main components: the trajectory planner and the trajectory follower. 
The former pre-computes optimal trajectories that connect two arbitrary states while minimizing the force exerted by the thrusters.
The trajectory follower computes continuous control actions to follow these and any other physically feasible trajectory using a \gls{tvlqr} formulation. 
To abide by the binary nature of the thrusters, it uses $\Sigma\Delta$-modulation to transfer the continuous force command onto the thrusters.
Finally, a \gls{kf} estimates the system state, providing feedback for the control architecture.

This work further develops a simulation of the overall system that takes measurement noise and the unevenness of the floor into account. 
When testing the control architecture in this simulation, the controller achieves at least \SI{16}{\centi\meter} average Euclidean and \SI{5}{\degree} average angular error.
The controller proves to be robust to arbitrary initial poses on the flat-floor.
In a Monte-Carlo simulation where the robot is spawned at arbitrary initial poses and tasked with finding and following optimal trajectories to the origin, the controller achieves a 100\% success rate.

The controller is further evaluated on the physical system. 
A straight-line trajectory is also followed successfully but experiences a drop in performance compared to the simulation. 
The average euclidean error approximately doubles, and the angular error increases by one order of magnitude.
The increase in error is mostly attributed to a lack of precise system identification and control authority.

A thorough system identification, including a more precise characterization of individual thrusters and their thrust vectors, will improve the system's performance in the future.
This improved model is helpful in two domains: the output trajectories of the planner will be more representative of the physical system, and the controller will follow these better.
Redesigning some system components to increase control authority will also help to improve accuracy.
The two most promising options are decreasing the overall weight and inertia of the system and increasing the reaction wheel inertia. 

Further work at the \gls{orl} intends to implement, test, and compare multiple implementations based on the software framework developed in this work.
More specifically, the future work will include improving the state estimation by adding additional sensors such as a \gls{imu} and eventually decoupling it from the global positioning system. 
One approach to improve the control will also be to solve the trajectory optimization problem online and control the system in a \gls{mpc} fashion, increasing resilience to disturbances.

\addtolength{\textheight}{-12cm}   





\bibliographystyle{ieeetr}
\bibliography{papers}

\end{document}